\documentclass[letterpaper, 10 pt, conference]{ieeeconf}  

\IEEEoverridecommandlockouts                              
\usepackage[utf8]{inputenc}
\usepackage[T1]{fontenc}
\usepackage{graphicx}
\usepackage{float} 
\usepackage{ragged2e}
\usepackage{booktabs}
\usepackage{bbding}
\usepackage{multirow}
\usepackage{makecell}
\usepackage{amsmath,amssymb,amsfonts}
\usepackage[table,xcdraw]{xcolor}
\usepackage{cite}
\usepackage{mathptmx} 
\usepackage{mathptmx} 
\usepackage[colorlinks,linkcolor=blue]{hyperref}
\usepackage{cleveref}
\usepackage{algorithmic}
\usepackage{algorithm}
\usepackage{tabularx}
\usepackage[caption=true,font=normalsize,labelfont=sf,textfont=sf]{subfig}
\usepackage{textcomp}
\usepackage{stfloats}

\hypersetup{
    colorlinks=true, 
    linkcolor=blue,  
    filecolor=blue,  
    urlcolor=black,   
    citecolor=blue,  
    pdfborder={0 0 0} 
}

\Crefname{figure}{Fig.}{Figs.}
\Crefname{equation}{Eq.}{Eqs.}

\def\eg{\emph{e.g.}}

\def\etal{\emph{et al.}}
\def\ie{\emph{i.e.}}

\newcommand{\verticaltext}[2][0pt]{%
  \raisebox{#1}{\parbox[t]{1em}{\rotatebox[origin=c]{90}{#2}}}%
}

\title{\LARGE \bf
FedRC: A Rapid-Converged Hierarchical Federated Learning Framework in Street Scene Semantic Understanding
}

\author{Wei-Bin Kou$^{1,2,3}$, Qingfeng Lin$^{1}$, Ming Tang$^{3}$, Shuai Wang$^{4}$, \\Guangxu Zhu$^{2,*}$, and Yik-Chung Wu$^{1,*}$
\thanks{This work has been accepted by 2024 IEEE/RSJ International Conference on Intelligent Robots and Systems (IROS). This work was supported in part by Funding ACK: National Natural Science Foundation of China (Grant No. 62371313), in part by Guangdong Basic and Applied Basic Research Foundation (Grant No. 2022A1515010109), in part by Shenzhen-Hong Kong-Macau Technology Research Programme (Type C) (Grant No. SGDX20230821091559018), in part by Longgang District Special Funds for Science and Technology Innovation (Grant No. LGKCSDPT2023002), and in part by the National Natural Science Foundation of China (Grant No. 62371444).}
\thanks{$^{*}$Corresponding author: Guangxu Zhu (gxzhu@sribd.cn) and Yik-Chung Wu (ycwu@eee.hku.hk).}
\thanks{$^{1}$Department of Electrical and Electronic Engineering, The University of Hong Kong, Hong Kong, China.}%
\thanks{$^{2}$Shenzhen Research Institute of Big Data, Shenzhen, China.}%
\thanks{$^{3}$Department of Computer Science and Engineering, Southern University of Science and Technology, Shenzhen, China.}%
\thanks{$^{4}$Shenzhen Institute of Advanced Technology, Chinese Academy of Sciences, Shenzhen, China.}%
}

\begin{document}

\maketitle
\thispagestyle{empty}
\pagestyle{empty}

\begin{abstract}
\underline{S}treet \underline{S}cene \underline{S}emantic \underline{U}nderstanding (denoted as TriSU) is a crucial but complex task for world-wide distributed autonomous driving (AD) vehicles (e.g., Tesla). Its inference model faces poor generalization issue due to inter-city domain-shift.  Hierarchical Federated Learning (HFL) offers a potential solution for improving TriSU model generalization, but suffers from slow convergence rate because of vehicles' surrounding heterogeneity across cities. Going beyond existing HFL works that have deficient capabilities in complex tasks, we propose a rapid-converged heterogeneous HFL framework (FedRC) to address the inter-city data heterogeneity and accelerate HFL model convergence rate. In our proposed FedRC framework, both single RGB image and RGB dataset are modelled as Gaussian distributions in HFL aggregation weight design. This approach not only differentiates each RGB sample instead of typically equalizing them, but also considers both data volume and statistical properties rather than simply taking data quantity into consideration. Extensive experiments on the TriSU task using across-city datasets demonstrate that FedRC converges faster than the state-of-the-art benchmark by 38.7\%, 37.5\%, 35.5\%, and 40.6\% in terms of mIoU, mPrecision, mRecall, and mF1, respectively. Furthermore, qualitative evaluations in the CARLA simulation environment confirm that the proposed FedRC framework delivers top-tier performance.
\end{abstract}

\section{INTRODUCTION}
\underline{\textbf{S}}treet \underline{\textbf{S}}cene \underline{\textbf{S}}emantic \underline{\textbf{U}}nderstanding (denoted as TriSU) is a crucial but complex task for globally distributed autonomous driving (AD) vehicles \cite{10416354,9981567}.  Recently, a number of new approaches \cite{10342110,10342254,10341639} for TriSU have been proposed, achieving impressive results. However, such TriSU methods typically face a challenge in generalization, even in relatively minor domain-shift \cite{muhammad2022vision}. This challenge becomes more pronounced when dealing with large inter-city domain-shift.

\begin{figure*}[t]
\centering 
\includegraphics[width=\linewidth, height=0.6\linewidth]{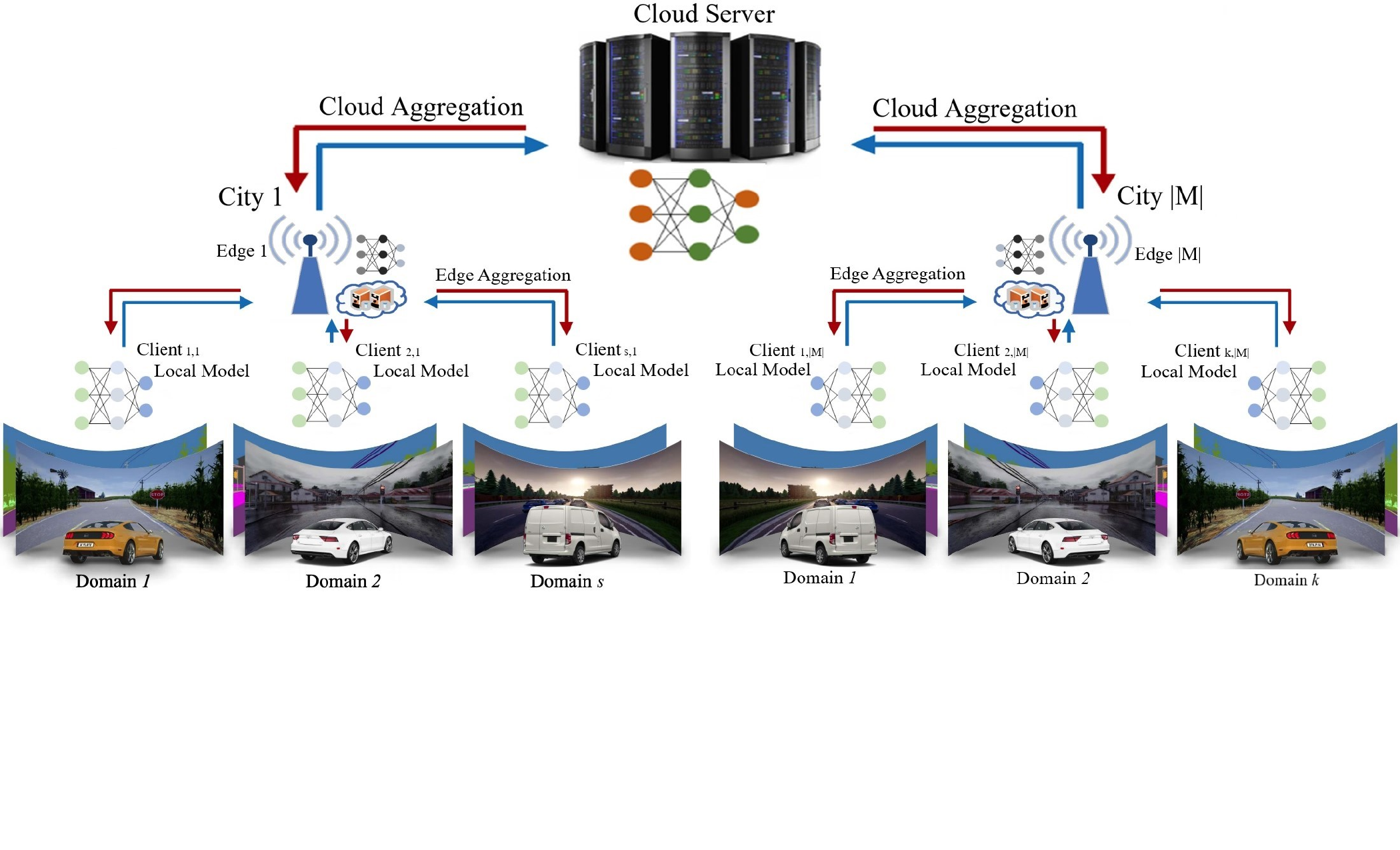}
\vspace{-3.8cm}
\captionsetup{justification=centering}
\caption{The illustration of Hierarchical Federated Learning (HFL) on TriSU task. \textbf{M} is the set of participating cities.
}
\label{Fig.HFL_TriSU}
\vspace{-0.4cm}
\end{figure*}

Hierarchical Federated Learning (HFL) \cite{liu2019clientedgecloud,10342134,wu2024hierarchical} (a variant of Federated Learning (FL) \cite{https://doi.org/10.48550/arxiv.1602.05629, feddrive2022,wang2022federated}), provides a promising framework not only to enhance TriSU model generalization in inter-city setting but also to improve communication efficiency.  Specifically, in the scenario considered in our work, we establish an edge server in each city. Within each city, all participating vehicles communicate their TriSU models with the edge server. We also set up a global cloud server that communicates models with all the edge servers. Our HFL setting on TriSU task is summarized in \Cref{Fig.HFL_TriSU}.

\begin{figure}[t]
\centering 
\includegraphics[width=\linewidth, height=0.5\linewidth]{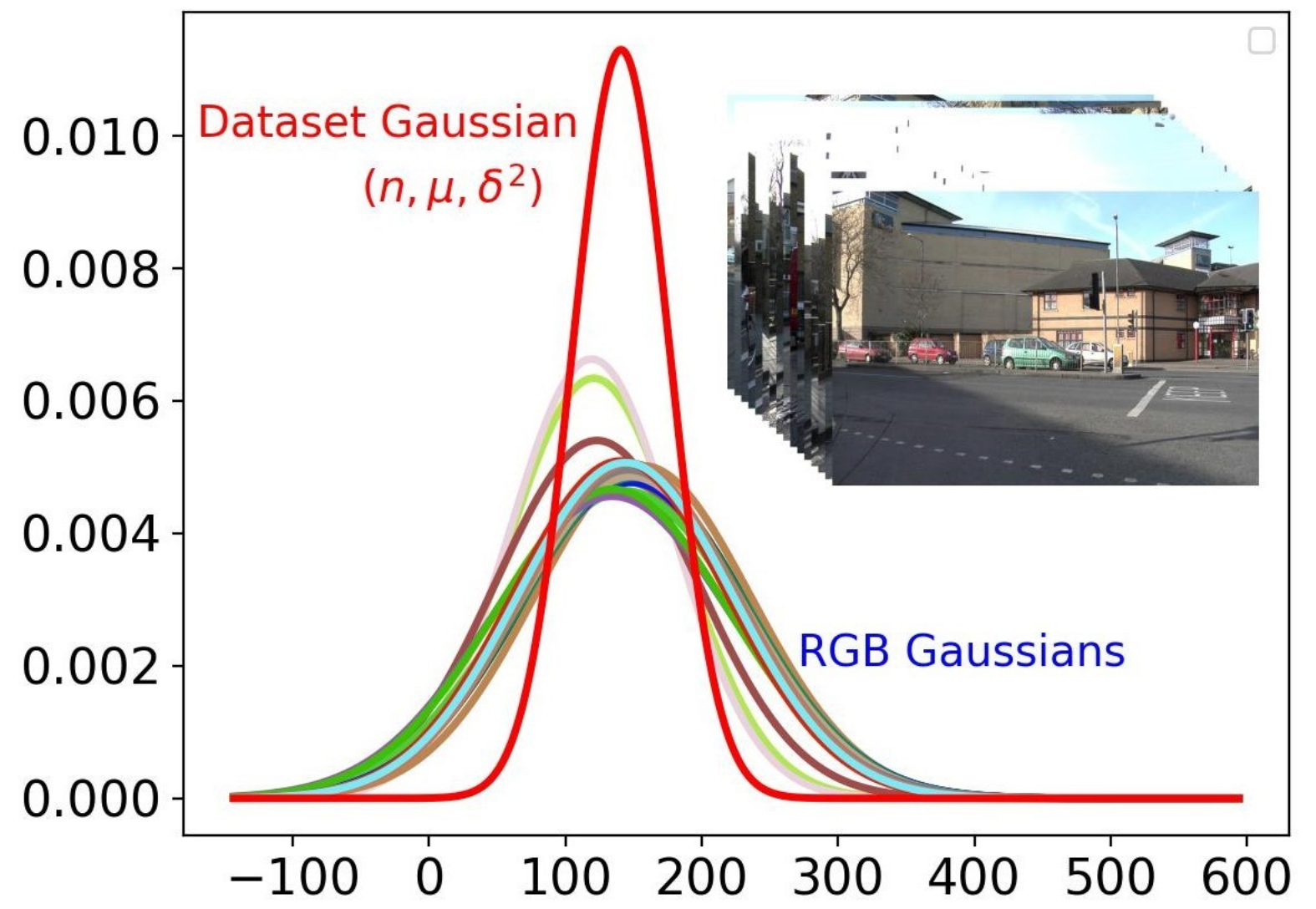}
\captionsetup{justification=centering}
\caption{The illustration of estimated RGB image Gaussians and RGB dataset Gaussian. $n, \mu, \delta^2$ represent the dataset size, mean and variance of dataset Gaussian distribution, respectively.
}
\label{Fig.dataset_gaussian}
\vspace{-0.7cm}
\end{figure}

HFL enhances TriSU model generalization by involving multiple $rounds$. In each $round$, HFL performs TriSU model learning in a two-stage process: (\textbf{I}) multiple edge aggregations followed by a (\textbf{I}) cloud aggregation. In edge aggregation stage, TriSU model aggregation at each edge server occurs through weighted averaging of all connected vehicles' models. The weight is typically defined as the $proportion$ of the vehicle's dataset size compared to the edge server's virtual dataset size.
In this stage, the aggregated model converges faster thanks to low data heterogeneity within one city, where $proportion$-based weight can approximately represent the vehicle's contribution in edge aggregation process. However, in cloud aggregation stage, the model converges slowly or even diverges due to large heterogeneity of far-away geographically distributed data from different cities. In this stage, the conventional $proportion$-based weight (\ie, the $proportion$ of the edge server's virtual dataset size compared to the cloud's virtual dataset size) has deficient ability to determine how much edge's model contributes in cloud aggregation, because it equalizes all samples and ignores the statistical distribution discrepancy among inter-city datasets, slowing down HFL model convergence. 
While some works \cite{10.1145/3594779} have proposed new kinds of weight instead of $proportion$-based weight fundamentally to accelerate model convergence by measuring data heterogeneity, their approaches have deficient capabilities to accelerate HFL model convergence in cloud aggregation stage in inter-city setting on complex TriSU task. \cite{10.1145/3594779} developed a kind of weight based on all vehicles' histograms, but it suffers from privacy leakage and consuming already stringent communication resource due to histograms transfer. 

In this paper, we propose FedRC to overcome HFL data heterogeneity and accelerate its convergence on TriSU task in inter-city setting. Specifically, our proposed FedRC is based on two points: \textbf{(I)} we model the distribution of each RGB image's pixel values as a Gaussian distribution, which can differentiate the contribution of each RGB sample from others instead of simply equalizing their contribution. \textbf{(II)} we further model RGB dataset using a Gaussian distribution by averaging all included RGB samples' Gaussian distributions, which considers both dataset size (\ie, $n$) and data statistics (\ie, $\mu, \delta^2$). These two points are illustrated in \Cref{Fig.dataset_gaussian}. 
Based on this, in the context of HFL, datasets on vehicles or covered by edge servers and cloud server are all modelled as Gaussian distributions which can be used to measure data heterogeneity.

To summarize, our main contributions are listed as follow:
\begin{itemize}
    \item To our knowledge, this is the first attempt to use Gaussian distributions to describe RGB images and datasets in HFL aggregation weight design for the TriSU task. This approach can handle inter-city data heterogeneity, because it not only values each RGB sample for its unique characteristics rather than treating all samples the same, but also considers both data quantity and statistical properties rather than solely considering data volume. 
    \item We propose a new method for assigning weights that uses statistical data to measure how much local and global datasets are related. This design is implemented in the HFL TriSU model to accelerate convergence by integrating data samples with greater similarity. This targeted approach to data integration facilitates more efficient learning and expedites the model's progression towards optimal performance. 
    \item Evaluation and experimental analysis are conducted on FedRC on TriSU task. The results show that FedRC does better than other top-performing methods on real-world data and on simulated data from CARLA \cite{dosovitskiy2017carla}.
\end{itemize}

\section{Related Work}
\label{related_work}
\subsection{Federated Learning (FL)}
FL is a decentralized and distributed machine learning paradigm that prioritizes data privacy preservation \cite{dong2022federated,karimireddy2020scaffold} and requires communication-efficient method \cite{10529194,lin2023communication} to reduce communication overheads and accelerate convergence.
For the initial FedAvg \cite{https://doi.org/10.48550/arxiv.1602.05629}, it aggregates vehicles' model parameters through weighted averaging at the server. However, some studies \cite{wang2021addressing,huang2021personalized} have found that data heterogeneity can negatively slow down convergence rate.
To address this issue, several strategies have been proposed \cite{li2020federated,acar2021federated}. For example, 
FedProx \cite{li2020federated} introduces a proximal regularization term on local models, ensuring updated local parameters remain close to the global model and preventing gradient divergence. 
FedDyn \cite{acar2021federated} uses a dynamic regularizer for each device to align global and local objectives.
Recently, personalized FL \cite{kou2024pfedlvm} is proposed to enhance the each client's model performance. 
However, these existing methods often underperform in complex tasks, such as object detection and semantic segmentation \cite{miao2023fedseg}. Although some existing works consider data heterogeneity from statistical prospective, they generally sacrifice data privacy because of involving transfer of raw data, histogram, etc\cite{10.1145/3594779, kou2023communication}.

\subsection{Street Scene Semantic Understanding (TriSU)}
TriSU is a field within computer vision and robotics focused on enabling machines to interpret and understand the content of street scenes, typically through various forms of sensory data such as images and videos. This capability is crucial for applications like autonomous driving\cite{10342102,10274109}. TriSU assigns a class label to every pixel in an image. This process is crucial for understanding the layout of the street scene, including the road, pedestrian, sidewalks, buildings, and other static and dynamic elements. Modern TriSU heavily relies on machine learning (ML), particularly deep learning (DL) techniques. Initially, Fully Convolutional Networks (FCNs)-based models significantly improve the performance of this task \cite{zhou2022rethinking}. In recent years, Transformer-based approaches \cite{10341519} have also been proposed for semantic segmentation. Recently, Bird's Eye View (BEV) \cite{9697426} technique is widely adopted for road scene understanding. 

\section{Methodology}
\label{methodology}
\subsection{HFL Formulation}

\begin{table}[t]
    \centering
    \renewcommand{\arraystretch}{1.0}
    \setlength{\tabcolsep}{15.0pt}
    \caption{Key Notations of HFL Formulation}
    \begin{tabularx}{\linewidth}{ll}
    \hline
        \textbf{Symbols} & \textbf{Definitions} \\ \hline
        $\mathop{e}$ & Edge server (Edge for short) ID \\
        $\mathop{\{c,e\}}$ & Vehicle ID \\ 
        $\mathcal{C}_e$ & Vehicle set connected to Edge $\mathop{e}$ \\ 
        $\mathcal{M}$ & Edge server set \\ 
        $\mathcal{D}_{c,e}$ & Training dataset on Vehicle $\mathop{\{c,e\}}$ \\  
        $\mathcal{D}_{e}$ & Training dataset virtually covered by Edge $\mathop{e}$ \\  
        $\mathcal{D}$ & Entire training dataset covered by Cloud \\ 
        $\mathcal{\omega}_{c,e}$ & Model parameters on Vehicle $\mathop{\{c,e\}}$ \\  
        $\mathcal{\omega}_{e}$ & Aggregated model parameters on Edge $\mathop{e}$ \\  
        $\mathcal{\omega}$ & Global aggregated model parameters on Cloud \\  
        $p_{c,e}$ & Aggregation weight for $\mathcal{\omega}_{c,e}$ \\  
        $p_{e}$ & Aggregation weight for $\mathcal{\omega}_{e}$\\  
        $\tau_1$ & Edge aggregation interval (EAI) \\ 
        $\tau_2$ & Cloud aggregation interval (CAI) \\
        $K$ & Total number of edge aggregation \\ 
        $R$ & Total number of cloud aggregation \\ \hline
    \end{tabularx}
\label{tab:HFRS}
\end{table}
The key notations in HFL are listed in \Cref{tab:HFRS}. We consider a HFL consisting of a cloud server, $\mathcal{|M|}$ edge servers and $ \sum_{e=1}^{\mathcal{|M|}} |\mathcal{C}_e|$ vehicles. Vehicle $\{c,e\}$ represents the $c$-th vehicle associated with Edge $e$, where $c = 1, 2, \cdots, |\mathcal{C}_e|$. Vehicle $\{c,e\}$ has a local dataset $\mathcal{D}_{c,e}$ with size $|\mathcal{D}_{c,e}|$. The Edge $e$ virtually covers dataset $\mathcal{D}_e \triangleq \cup_{c=1}^{|\mathcal{C}_e|} \mathcal{D}_{c,e}$ with size $|\mathcal{D}_e|$. Similarly, the cloud server virtually covers dataset $\mathcal{D} \triangleq \cup_{e=1}^{|\mathcal{M}|} \mathcal{D}_{e}$ with size $|\mathcal{D}|$.

\subsubsection{Vehicle Training}
In local update $u$ (refers to index of local iteration), Vehicle $\{c,e\}$ trains its local model $\mathbf{\omega}_{c,e}$ based on dataset $\mathcal{D}_{c,e}$. We define loss function of $j$-th sample out of $\mathcal{D}_{c,e}$ as $\mathcal{E}(\mathbf{\omega}_{c,e},\mathcal {D}_{c,e}^{(j)})$, and the training is given by 
\begin{equation}
\label{equa:robot}
\mathop{\mathrm{min}}_{\mathbf{\omega}_{c,e}}~\mathcal{L}_{c,e}(\mathcal{\omega}_{c,e}) =
\frac{1}{|\mathcal{D}_{c,e}|}
\sum_{\mathcal{D}_{c,e}^{(j)}\in\mathcal{D}_{c,e}
}\mathcal{E}(\mathbf{\omega}_{c,e},\mathcal{D}_{c,e}^{(j)}).
\end{equation}

\subsubsection{Edge Aggregation}
When vehicle local update $u$ = $k\tau_1$, $k=1,2,\cdots, K$, each edge server receives vehicles' models every $\tau_1$ local iterations and then performs edge aggregation:
\begin{align}
    \omega_e = \sum_{c=1}^{|\mathcal{C}_e|} p_{c,e}\omega_{c,e}, ~~~
    \mathcal{L}_{e}(\omega_e) = \sum_{c=1}^{|\mathcal{C}_e|} p_{c,e}\mathcal{L}_{c,e}(\omega_e).
\end{align}

\subsubsection{Cloud Aggregation}
When vehicle local update $u$ = $r\tau_1\tau_2$, $r=1,2,\cdots, R$, the cloud server receives models from all edge servers every $\tau_2$ edge aggregations and performs cloud aggregation:
\begin{align}
    \omega = \sum_{e=1}^{|\mathcal{M}|} p_e\omega_{e}, ~~~ 
    \mathcal{L}(\omega) = \sum_{e=1}^{|\mathcal{M}|} p_e\mathcal{L}_{e}(\omega).
\end{align}
Then the cloud will redistribute the aggregated model $\omega$ to all edge servers and then to all vehicles. Our goal is to minimize the global loss $\mathcal{L}(\omega)$ of HFL, such that the global model $\omega$ is the weighted average of all vehicles' model:
\begin{align}
\mathop{\mathrm{min}}_{\mathbf{\omega} \in \mathbb{R}^d} \mathop{\mathcal{L}(\mathbf{\omega})} 
& \triangleq \sum_{e=1}^{|\mathcal{M}|} p_e\mathcal{L}_{e}(\mathbf{\omega})
= \sum_{e=1}^{|\mathcal{M}|}p_e\sum_{c=1}^{|\mathcal{C}_e|} p_{c,e}\mathcal{L}_{c,e}(\mathbf{\omega}),
\nonumber \\
\mathrm{s.t.} ~ \omega & = \sum_{e=1}^{|\mathcal{M}|}p_e\sum_{c=1}^{|\mathcal{C}_e|}p_{c,e} \omega_{c,e}.
\label{equal:optimal}
\end{align}

\subsection{FedRC Framework}
In this section, we will introduce the proposed FedRC. The mathematical principle of the FedRC framework comes from FL convergence analysis\cite{wang2023batch}. Wang \etal \cite{wang2023batch} reports that the slow convergence rate can be attributed to the statistical discrepancy between local datasets and the global dataset, especially for a non-i.i.d. setting. Precisely, in \Cref{equal:optimal}, the typical FL weights:
\begin{align}
    p_{c,e} = \frac{|\mathcal{D}_{c,e}|}{|\mathcal{D}_{e}|},~~~ 
    p_e = \frac{|\mathcal{D}_{e}|}{|\mathcal{D}|}, 
\end{align}
treat that each RGB sample contributes equally in aggregation. Such weight design fails to underscore the statistical discrepancy between local datasets and global dataset.

Motivated by this, we propose FedRC to measure this statistical discrepancy and then to further accelerate HFL convergence rate in inter-city (non-i.i.d) setting. Our observations indicate that the distribution of pixel intensities in RGB images (or individual channels of color images) displays a bell-curve shape when visualized as a histogram, which is a characteristic feature of a Gaussian distribution. Therefore, in the proposed FedRC framework, pixel value's distribution of both individual RGB images and entire RGB datasets are modelled as Gaussian distributions. Based on such points, we detail the proposed FedRC in the following progressive steps:

\subsubsection{Step I: Distribution Estimation of Single RGB Image}
\begin{figure}[tp]
\centering 
\vspace{-0.3cm}
\includegraphics[width=\linewidth]{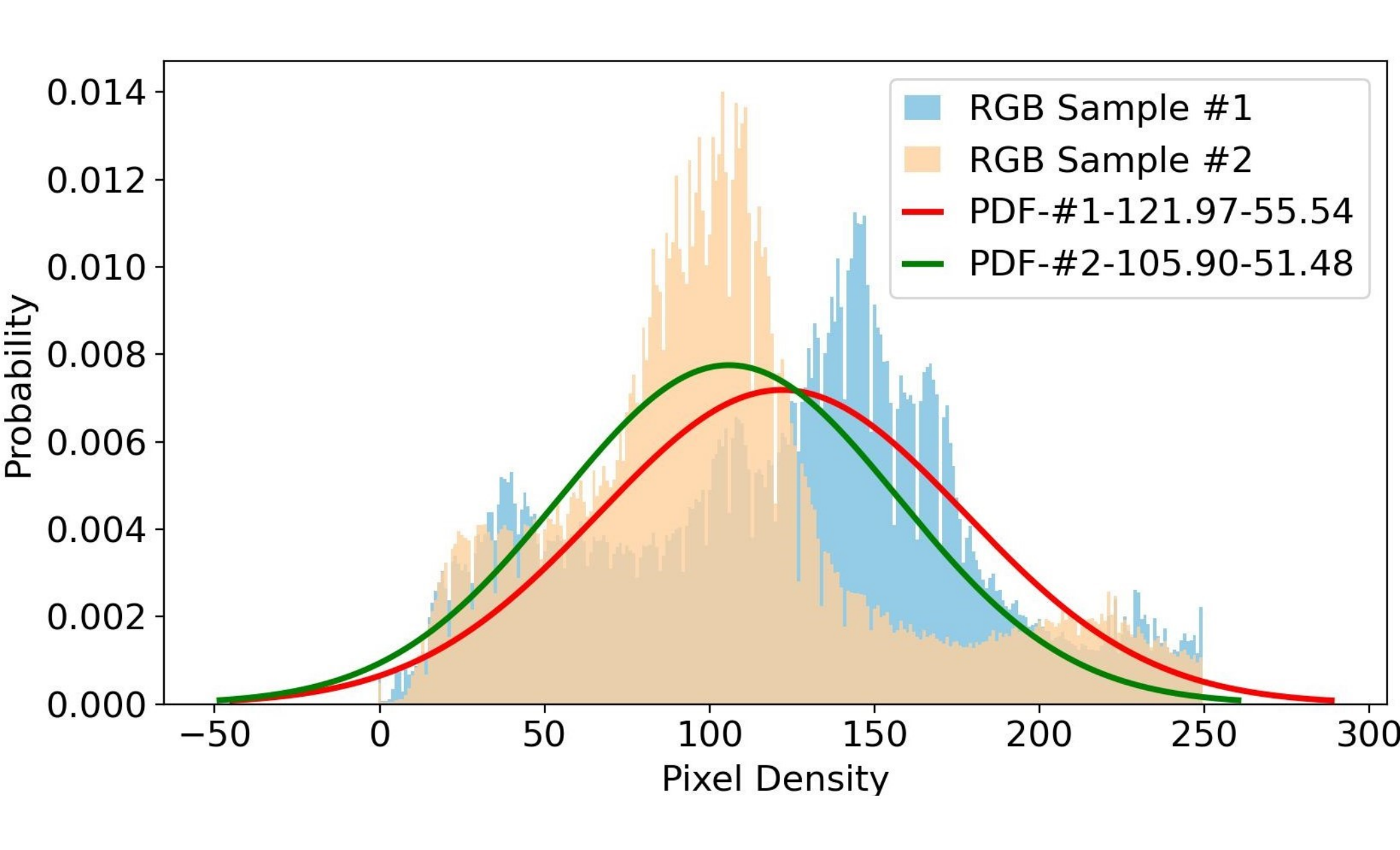}
\vspace{-0.8cm}
\caption{This figure illustrates the normalized histogram and probability density function (PDF) of two RGB samples. For example, with respect to ``RGB Sample \#1'', the estimated mean and variance of Gaussian distribution are 121.97 and 55.54, respectively.}
\label{Fig.2_rgb_comp}
\vspace{-0.4cm}
\end{figure}

For single RGB image with the resolution $\mathcal{W} \times \mathcal{H}$, we suppose the pixel value $\mathcal{X}_i$ is a Gaussian random variable, \ie, $\mathcal{X}_i \sim \mathcal{N}(\mu_i, \delta^2_i)$. The $\mu_i$ and $\delta^2_i$ can be estimated using total $L = 3 \times \mathcal{W} \times \mathcal{H}$ samples according to \Cref{eq:rgb_delta}:
\begin{align}
   \mu_i = \frac{1}{L}\sum_{l=1}^L x_l, ~~~
   \delta^2_i = \frac{1}{L-1} \sum_{l=1}^L (x_l - \mu_i)^2,
   \label{eq:rgb_delta}
\end{align}
where $x_l$ means one pixel value from the RGB image. \Cref{Fig.2_rgb_comp} presents two estimated examples of RGB image.

\subsubsection{Step II: RGB Dataset Distribution Estimation of Vehicles, Edge Servers and Cloud Server}
For Vehicle $\{c,e\}$, its dataset $\mathcal{D}_{c,e}$ contains $n_{c,e}$ (equals to $|\mathcal{D}_{c,e}|$) RGB images. Based on \textit{Step I}, we can model the $i$-th ($1 \leq i \leq n_{c,e}$) image as $\mathcal{X}_i \sim \mathcal{N}(\mu_i, \delta_i^2)$. Furthermore, we define the Gaussian distribution of $\mathcal{D}_{c,e}$ is $\mathcal{X}_{c,e} = 1 / n_{c,e} \sum_{i=1}^{n_{c,e}} \mathcal{X}_i \sim \mathcal{N}(\mu_{c,e}, \delta_{c,e}^2)$, where $\mu_{c,e}$ and $\delta_{c,e}^2$ can be estimated by \Cref{eq:dataset_delta}: 
\begin{align}
    n_{c,e} = |\mathcal{D}_{c,e}|,~
    \mu_{c,e} = \frac{1}{n_{c,e}} \sum_{i=1}^{n_{c,e}} \mu_i,~  
    \delta_{c,e}^2 = \frac{1}{n_{c,e}^2} \sum_{i=1}^{n_{c,e}} \delta_i^2.
\label{eq:dataset_delta}
\end{align}
Taking the dataset size $n_{c,e}$ into consideration, we can use a three-element tuple $(n_{c,e}, \mu_{c,e}, \delta_{c,e}^2)$ to represent $\mathcal{D}_{c,e}$.

For the Edge $e$, it receives $(n_{c,e}, \mu_{c,e}, \delta_{c,e}^2)$ from all connected vehicles. Then Edge $e$ can calculate its own Gaussian distribution parameters by \Cref{eq:edge_delta}:
\begin{align}
    n_{e} = \sum_{c=1}^{|\mathcal{C}_e|} n_{c,e}, ~
    \mu_{e} = \frac{1}{n_{e}} \sum_{c=1}^{|\mathcal{C}_e|} n_{c,e}\mu_{c,e}, ~ 
    \delta_{e}^2 = \frac{1}{n_{e}^2} \sum_{c=1}^{|\mathcal{C}_e|} n_{c,e}^2\delta_{c,e}^2.
    \label{eq:edge_delta}
\end{align}

Similarly, for the Cloud, it receives three-element tuple $(n_e, \mu_{e}, \delta_{e}^2)$ from all edge servers, and then can calculate its own Gaussian distribution parameters based on \Cref{eq:cloud_delta}:
\begin{align}
    n = \sum_{e=1}^{|\mathcal{M}|} n_{e}, ~ 
    \mu = \frac{1}{n} \sum_{e=1}^{|\mathcal{M}|} n_{e}\mu_{e}, ~
    \delta^2 = \frac{1}{n^2} \sum_{e=1}^{|\mathcal{M}|} n_{e}^2\delta_{e}^2.
    \label{eq:cloud_delta}
\end{align}

\begin{figure*}[tp]
\centering
\subfloat[Edge1]{\includegraphics[width=0.24\linewidth,height=0.15\linewidth]{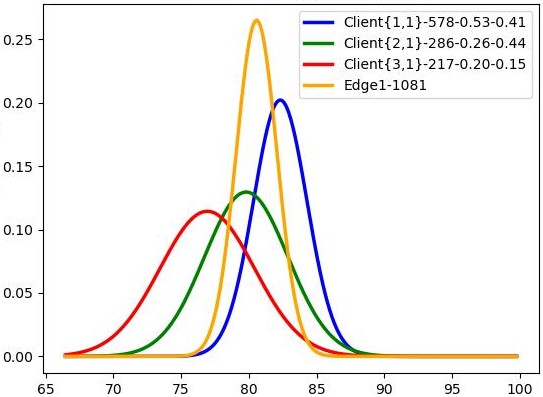}
\label{Fig.city_edge1}
}
\subfloat[Edge2]{\includegraphics[width=0.24\linewidth,height=0.15\linewidth]{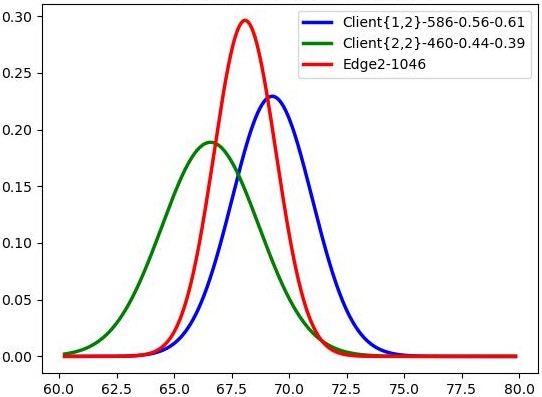}
\label{Fig.city_edge2}
}
\subfloat[Edge3]{\includegraphics[width=0.24\linewidth,height=0.15\linewidth]{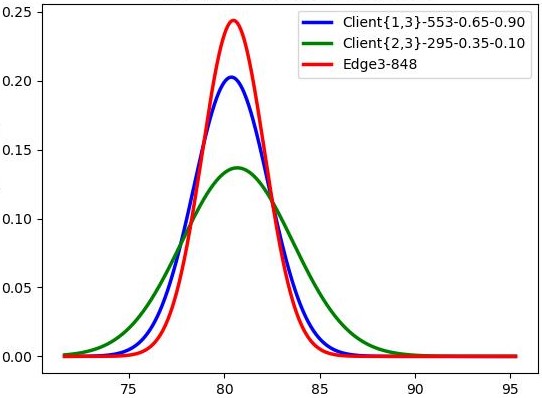}
\label{Fig.city_edge3}
}
\subfloat[Cloud]{\includegraphics[width=0.24\linewidth,height=0.15\linewidth]{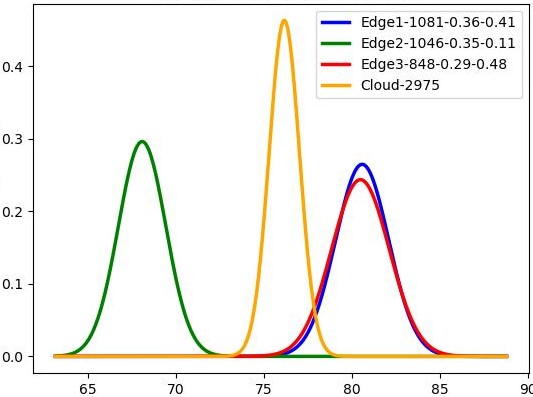}
\label{Fig.city_cloud}
}
\caption{FedRC result. The legend $'Client\{1,1\}-578-0.53-0.41'$ in \Cref{Fig.city_edge1} can be separated into four parts by $'-'$. They represent vehicle ID, dataset size, $proportion$-based weight and FedRC weight, respectively. The legend $'Edge1-1081'$ in \Cref{Fig.city_edge1} means Edge $1$ has virtual dataset with 1081 size. The legends in \Cref{Fig.city_edge2,Fig.city_edge3,Fig.city_cloud} share the similar meaning with \Cref{Fig.city_edge1}. It is observed that FedRC weights are better than $proportion$-based weight for aggregation. For example, in the \Cref{Fig.city_cloud}, the Edge 2 distribution is far away from the Cloud distribution, it should have a smaller weight for model aggregation, which FedRC weight fits whereas $proportion$-based weight does not.}
\label{Fig.dataset_dist1}
\vspace{-0.3cm}
\end{figure*}

\subsubsection{Step III: Distance between Local and Global Dataset}
Given two Gaussian distributions $\mathcal{D}_1 \sim \mathcal{N}(\mu_{D_1}, \delta_{D_1}^2)$ and $\mathcal{D}_2 \sim \mathcal{N}(\mu_{D_2}, \delta_{D_2}^2)$, we propose using Bhattacharyya distance (BD) \cite{bhattacharyya1943measure} termed $D_B(\mathcal{D}_1, \mathcal{D}_2)$ to measure the distance between them. BD can be calcuated by \Cref{eq:bc_def,eq:pdfs,eq:BD}: 

\begin{equation}
BC(\mathcal{D}_1, \mathcal{D}_2) = \int \sqrt{f_1(x) f_2(x)} \, dx,
\label{eq:bc_def}
\end{equation}
\begin{equation}
f_i(x) = \frac{1}{\sqrt{2\pi}\delta_{D_i}} \exp\left(-\frac{(x - \mu_{D_i})^2}{2\delta_{D_i}^2}\right), \quad i=1,2,
\label{eq:pdfs}
\end{equation}

\begin{equation}
D_B(\mathcal{D}_1, \mathcal{D}_2) = -\ln(BC(\mathcal{D}_1, \mathcal{D}_2)).
\label{eq:BD}
\end{equation}
The BD can be formulated finally as following \Cref{eq:bd}:
\begin{equation}
D_B(\mathcal{D}_1, \mathcal{D}_2) = \frac{1}{4} \frac{(\mu_{D_1} - \mu_{D_2})^2}{\delta_{D_1}^2 + \delta_{D_2}^2} + \frac{1}{2} \ln\left(\frac{\delta_{D_1}^2 + \delta_{D_2}^2}{2 \delta_{D_1} \delta_{D_2}}\right),
\label{eq:bd}
\end{equation}
where the first term indicates the divergence between the two distributions, while the subsequent term underscores the disparity in the distribution's dispersion. The primary benefit of the BD is its consideration of the full distribution, rather than merely its mean and variance. This attribute renders it particularly apt for datasets with considerable variability.

On top of \Cref{eq:bd}, we can calculate the distance between Vehicle $\{c,e\}$ and Edge $e$ by $D_B(\mathcal{D}_{c,e}, \mathcal{D}_e)$, and distance between Edge $e$ and Cloud by $D_B(\mathcal{D}_{e}, \mathcal{D})$.

\subsubsection{Step IV: FedRC Weights Calculation}
Based on distances $D_B(\mathcal{D}_{c,e}, \mathcal{D}_e)$ and $D_B(\mathcal{D}_{e}, \mathcal{D})$ in \textit{Step III},  $p_{c,e}$ and $p_e$ can be computed as \Cref{eq:coef}:
\begin{align}
    p_{c,e} = \frac{1 / D_B(\mathcal{D}_{c,e}, \mathcal{D}_e)}{\sum_{c} (1 / D_B(\mathcal{D}_{c,e}, \mathcal{D}_e))},
    p_{e} = \frac{1 / D_B(\mathcal{D}_{e}, \mathcal{D})}{\sum_{e} (1 / D_B(\mathcal{D}_{e}, \mathcal{D}))}, 
\label{eq:coef}
\end{align}
which implies that the closer distance yields higher aggregation weight. When compared with $proportion$-based weight to equalize all RGB samples, the proposed approach can leverage personalized Gaussian distribution of each sample to accelerate HFL convergence on TriSU task.

In summary, we formulate FedRC in \Cref{alg:HierFedRC} (overall framework) and \Cref{alg:FedRC} (basic operation unit). Furthermore, we visualize the FedRC results as shown in \Cref{Fig.dataset_dist1}. 

\subsection{Complexity Analysis}
\subsubsection{Space Complexity}
In terms of space complexity, the storage demands are as follows: for $n$ RGB images, the space required is $2n$ units; for $|\mathcal{V}|$ vehicles, it is $3|\mathcal{V}|$ units; for $|\mathcal{M}|$ edge servers, it is $3|\mathcal{M}|$ units; and for the cloud server, 3 units are required. Thus, the total space requirement termed $S_{c, FedRC}$ for the FedRC system is expressed by \Cref{eq:fedgau_space_comp}:
\begin{align}
S_{c, FedRC} = 2n + 3|\mathcal{V}| + 3|\mathcal{M}| + 3.
\label{eq:fedgau_space_comp}
\end{align}
Under typical conditions where $n$ significantly exceeds $|\mathcal{V}|$ and $|\mathcal{M}|$ (\ie, $n \gg |\mathcal{V}|$ and $n \gg |\mathcal{M}|$), we can approximate the total space requirement $S_{c, FedRC}$ to be roughly $2n$, with the space complexity being O($n$).

\subsubsection{Time Complexity}
With regard to time complexity, we assume that the basic summation operation take the time of $t_p$. Therefore, the overall computation time for processing all RGB images is $6n\mathcal{W}\mathcal{H}t_p$; for all vehicles, it is $2nt_p$; for all edge servers, it is $3|\mathcal{V}|t_p$; and for the cloud server, it is $3|\mathcal{M}|t_p$. The cumulative time requirement $T_{c, FedRC}$ for the FedRC system is thus given by \Cref{eq:fedgau_time_comp}:
\begin{align}
T_{c, FedRC} = 6n\mathcal{W}\mathcal{H}t_p + 2nt_p + 3|\mathcal{V}|t_p + 3|\mathcal{M}|t_p.
\label{eq:fedgau_time_comp}
\end{align}
Considering that $n$ is much larger than $|\mathcal{V}|$ and $|\mathcal{M}|$ (\ie, $n \gg |\mathcal{V}|$ and $n \gg |\mathcal{M}|$). Moreover, given that the aspect ratio of an RGB image is generally denoted as $\alpha = \mathcal{W} / \mathcal{H}$ and the term $3\mathcal{W}\mathcal{H}$ is typically much greater than 1. $T_{c, FedRC}$ simplifies to the approximation shown in \Cref{eq:approx_fedgau_time_comp_simp}:
\begin{align}
T_{c, FedRC} \approx 6\alpha n\mathcal{H}^2t_p.
\label{eq:approx_fedgau_time_comp_simp}
\end{align}
Given this simplification, it becomes apparent that the total computation time is predominantly influenced by the number of images $n$ and the square of the height dimension $\mathcal{H}$ of the images. Thus, the time complexity of FedRC can be denoted as \(\text{O}(n\mathcal{H}^2)\).

\begin{algorithm}[t]
\caption{FedRC}
\begin{algorithmic}[1]
\STATE \textbf{Input:} Cloud server: \textbf{Cloud}, Edge set: $\mathbf{\mathcal{M}}$, Vehicle set: $\cup_{e=1}^{|\mathcal{M}|}\mathcal{C}_e$ \\
\STATE \textbf{Output:} Aggregation Weights: $\mathcal{P}$ \\
\STATE $Algo\ FedRC(\mathbf{Cloud}, \mathcal{M}, \cup_{e=1}^{|\mathcal{M}|}\mathcal{C}_e)$
\STATE \textbf{Edge Server Side:}
\FOR{$\mathbf{Edge}\ e\ in\ \mathbf{\mathcal{M}}$}
    \STATE $FedRC\_Base(\mathbf{Edge}\ e, \mathcal{C}_e)$  ~//Algorithm 2
\ENDFOR
\STATE ~\\
\STATE \textbf{Cloud Side:}
\STATE $FedRC\_Base(\mathbf{Cloud}, \mathcal{M})$ ~//Algorithm 2 
\end{algorithmic}
\label{alg:HierFedRC}
\end{algorithm}

\setlength{\textfloatsep}{8pt}
\begin{algorithm}[t]
\caption{FedRC\_Base}
\begin{algorithmic}[1]
\STATE \textbf{Input:} One server: $\mathbf{Server}$, Connected node set: $\mathbf{NS}$ \\
\STATE \textbf{Output:} Aggregation Weights: $\mathcal{P}$ \\
\STATE $Algo\ FedRC\_Base(\mathbf{Server}, \mathbf{NS}):$
\STATE \textbf{Node Side:}
\FOR{$\mathbf{Node}\ \mathcal{S}\ in\ \mathbf{NS}$}
    \STATE $n_\mathcal{S}\ \mu_\mathcal{S}, \delta^2_\mathcal{S} \leftarrow \Cref{eq:dataset_delta}$
    \STATE $Send\ n_\mathcal{S}\ \mu_\mathcal{S}, \delta^2_\mathcal{S} \Rightarrow Server$ \\
\ENDFOR
\STATE ~\\
\STATE \textbf{Server Side:}
\STATE $n, \mu, \delta^2 \leftarrow \Cref{eq:edge_delta}\ or\ \Cref{eq:cloud_delta}$
\FOR{$\mathbf{Node}\ \mathcal{S}\ in\ \mathbf{NS}$}
    \STATE $\mathcal{P}_{\mathcal{S}} = D_B((n_\mathcal{S}, \mu_{\mathcal{S}}, \delta^2_{\mathcal{S}}), (n, \mu, \delta^2))$ \\
\ENDFOR
\end{algorithmic}
\label{alg:FedRC}
\end{algorithm}  

\section{Experiments}
\label{experiments}
This section details experiments undertaken on the TriSU task across various cities. We aim to measure the acceleration of convergence and the enhancement of performance attributable to FedRC, employing metrics that are widely recognized and accepted.

\begin{table}[tp]
\vspace{-0.3cm}
    \centering
    \renewcommand{\arraystretch}{1.0}
    \setlength{\tabcolsep}{15.0pt}
    \caption{Hardware/Software configurations}
    \begin{tabularx}{\linewidth}{ll}
    \hline
        \textbf{Items} & \textbf{Configurations} \\ \hline
        CPU  & AMD Ryzen 9 3900X 12-Core \\ 
        GPU  & NVIDIA GeForce 3090 $\times$ 2\\ 
        RAM  & DDR4 32G \\ 
        DL Framework  & PyTorch @ 1.13.0+cu116 \\ 
        GPU Driver  & 470.161.03 \\ 
        CUDA  & 11.4 \\ 
        cuDNN  & 8302 \\ \hline
    \end{tabularx}
\label{Tab:configs}
\end{table}

\begin{table}[tp]
\vspace{-0.3cm}
    \centering
    \renewcommand{\arraystretch}{1.0}
    \setlength{\tabcolsep}{15.0pt}
    \caption{Training configurations}
    \begin{tabularx}{\linewidth}{ll}
    \hline
        \textbf{Items} & \textbf{Configurations} \\ \hline
        Loss  & nn.CrossEntropyLoss \\ 
        Optimizer  & nn.Adam \\ 
        Adam Betas  & (0.9, 0.999) \\ 
        Weight Decay  & 1e-4 \\ 
        Batch Size  & 8 \\ 
        Learning Rate  & 3e-4 \\ 
        \multirow{1}{*}{DNN Models}  & DeepLabv3+ \cite{chen2018encoderdecoder} \\
        \multirow{3}{*}{FL Algorithms} &FedAvg \cite{https://doi.org/10.48550/arxiv.1602.05629}, FedProx \cite{li2020federated}, FedDyn \cite{acar2021federated} \\
        ~ &FedAvgM \cite{hsu2019measuring}, FedIR \cite{hsu2020federated}, FedNova \cite{wang2020tackling} \\
        ~ &SCAFFOLD \cite{karimireddy2020scaffold} \\
        \hline
    \end{tabularx}
\label{Tab:train}
\end{table}

\subsection{Datasets, Evaluation Metrics and Implementation}
\subsubsection{Datasets}
The \textbf{Cityscapes} dataset \cite{Cordts2016Cityscapes} includes 2,975 training and 500 validation images with masks. The \textbf{Cityscapes} dataset includes 19 semantic classes, including vehicles, pedestrians and so forth. The training dataset is split into parts for HFL vehicles. The \textbf{CamVid} dataset \cite{brostow2008segmentation} totally includes 701 samples with 11 semantic classes. In our experiments, we split random-selected 600 samples into parts for HFL vehicles. The remaining 101 samples are used as test dataset. In addition, we will also implement FedRC on CARLA \cite{dosovitskiy2017carla} simulation platform to verify it qualitatively.

\subsubsection{Evaluation Metrics}
We evaluate our proposals on TriSU task using four metrics: \textbf{mIoU}, \textbf{mPrecision (mPre for short)}, \textbf{mRecall (mRec for short)}, and \textbf{mF1}. These metrics are defined as follows: 
\vspace{-0.2cm}
\begin{align}
    &mIoU = \frac{1}{\mathcal{C}}\sum_{c=1}^{\mathcal{C}}IoU_c = \frac{1}{\mathcal{C} * \mathcal{N}}\sum_{c=1}^{\mathcal{C}} \sum_{n=1}^{\mathcal{N}} \frac{TP_{n, c}}{FP_{n, c} + TP_{n, c} + FN_{n, c}},
    \nonumber
    \\
    \vspace{-0.2cm}
    &mPre = \frac{1}{\mathcal{C}}\sum_{c=1}^{\mathcal{C}}Pre_c = \frac{1}{\mathcal{C} * \mathcal{N}}\sum_{c=1}^{\mathcal{C}} \sum_{n=1}^{\mathcal{N}} \frac{TP_{n, c}}{FP_{n, c} + TP_{n, c}},
    \nonumber
    \\
    \vspace{-0.2cm}
    &mRec = \frac{1}{\mathcal{C}}\sum_{c=1}^{\mathcal{C}}Rec_c = \frac{1}{\mathcal{C} * \mathcal{N}}\sum_{c=1}^{\mathcal{C}} \sum_{n=1}^{\mathcal{N}} \frac{TP_{n, c}}{TP_{n, c} + FN_{n, c}},
    \nonumber
    \\
    \vspace{-0.2cm}
    &mF1 = \frac{1}{\mathcal{C}}\sum_{c=1}^{\mathcal{C}}F1_c = \frac{1}{\mathcal{C}}\sum_{c=1}^{\mathcal{C}} \frac{2 * Pre_c * Rec_c}{Pre_c + Rec_c},
\end{align}
where $TP$, $FP$, $TN$ and $FN$ are short for True Positive, False Positive, True Negative and False Negative, respectively. $\mathcal{C}$ denotes the number of semantic classes within the test dataset, with values set to 19 for the Cityscapes dataset and 11 for the CamVid dataset. Similarly, $\mathcal{N}$ signifies the size of the test dataset, which amounts to 500 for Cityscapes and 101 for CamVid.

\subsubsection{Implementation Details}
The main hardware and software configurations are listed in \Cref{Tab:configs}. The main training details are listed in \Cref{Tab:train}. Our experiments involve a comparison between the proposed FedRC and other several FL algorithms. Among these benchmarks, FedDyn, FedProx, and FedAvgM each include a hyperparameter which is set in brackets for notation, \eg, FedDyn(0.01). 

\begin{figure*}[tp]
\centering
\subfloat[mIoU]{\includegraphics[width=0.24\linewidth]{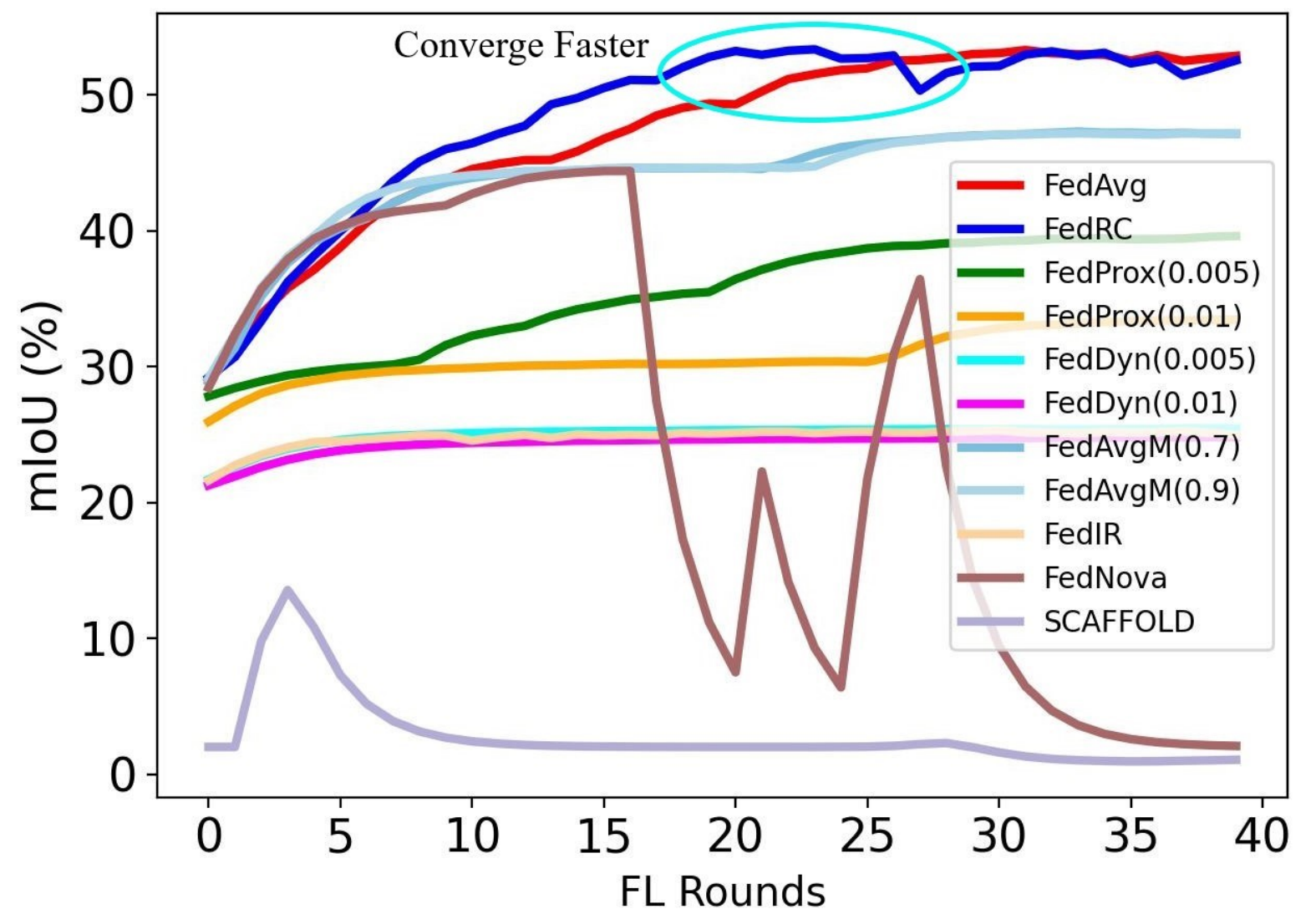}
\label{Fig.Metrics_mIoU}
}
\subfloat[mPrecision]{\includegraphics[width=0.24\linewidth]{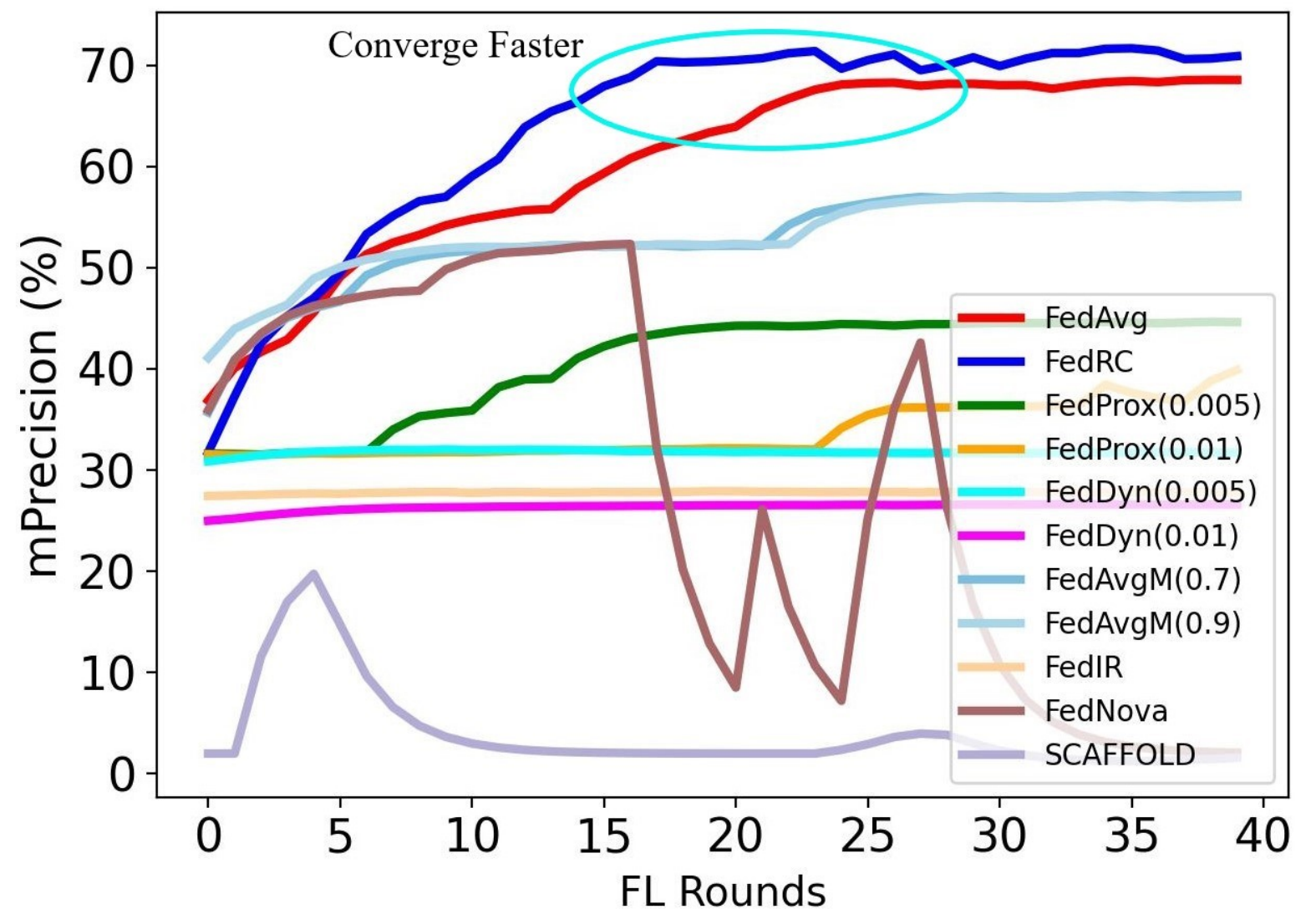}
\label{Fig.Metrics_mPre}
}
\subfloat[mRecall]{\includegraphics[width=0.24\linewidth]{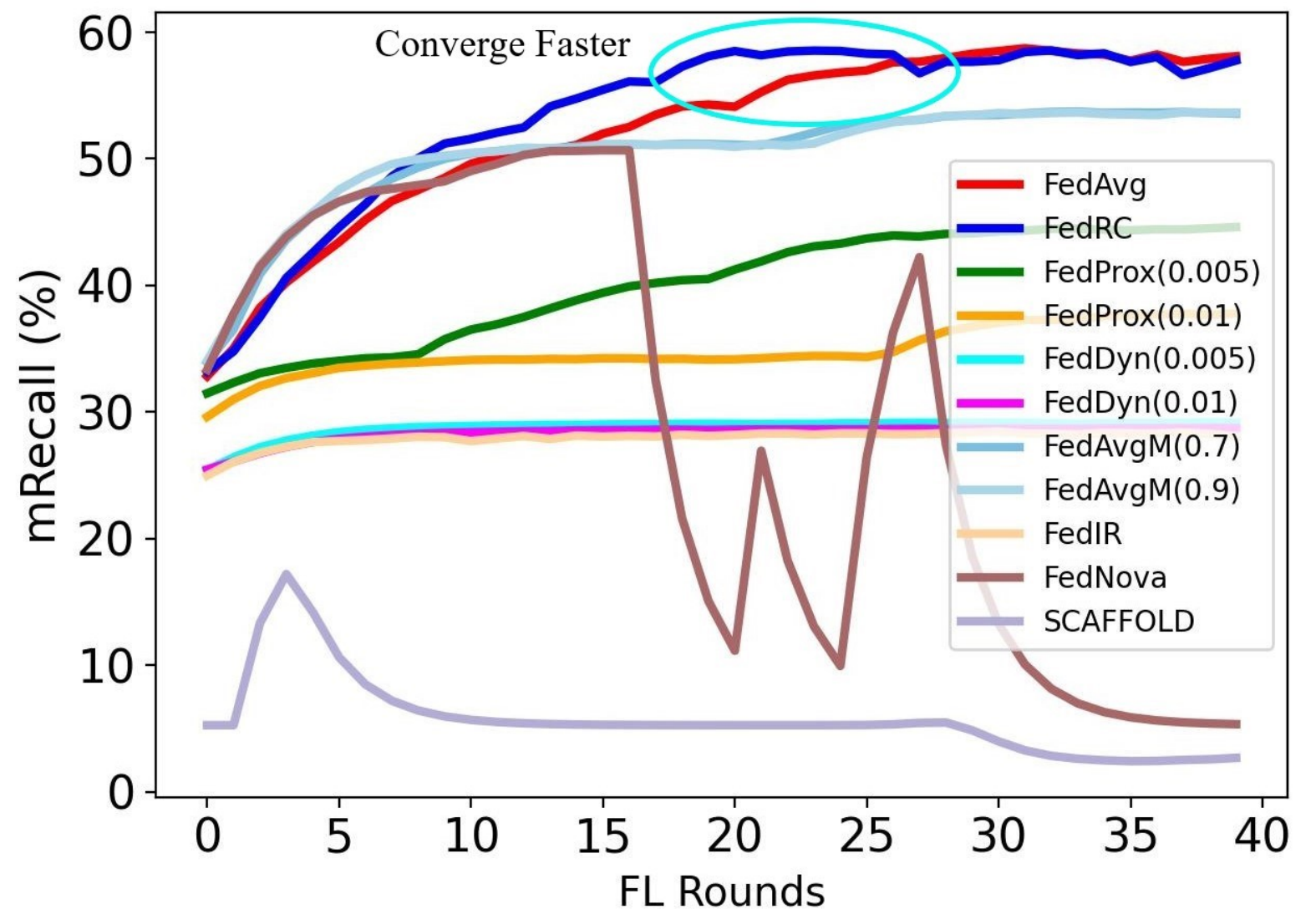}
\label{Fig.Metrics_mRec}
}
\subfloat[mF1]{\includegraphics[width=0.24\linewidth]{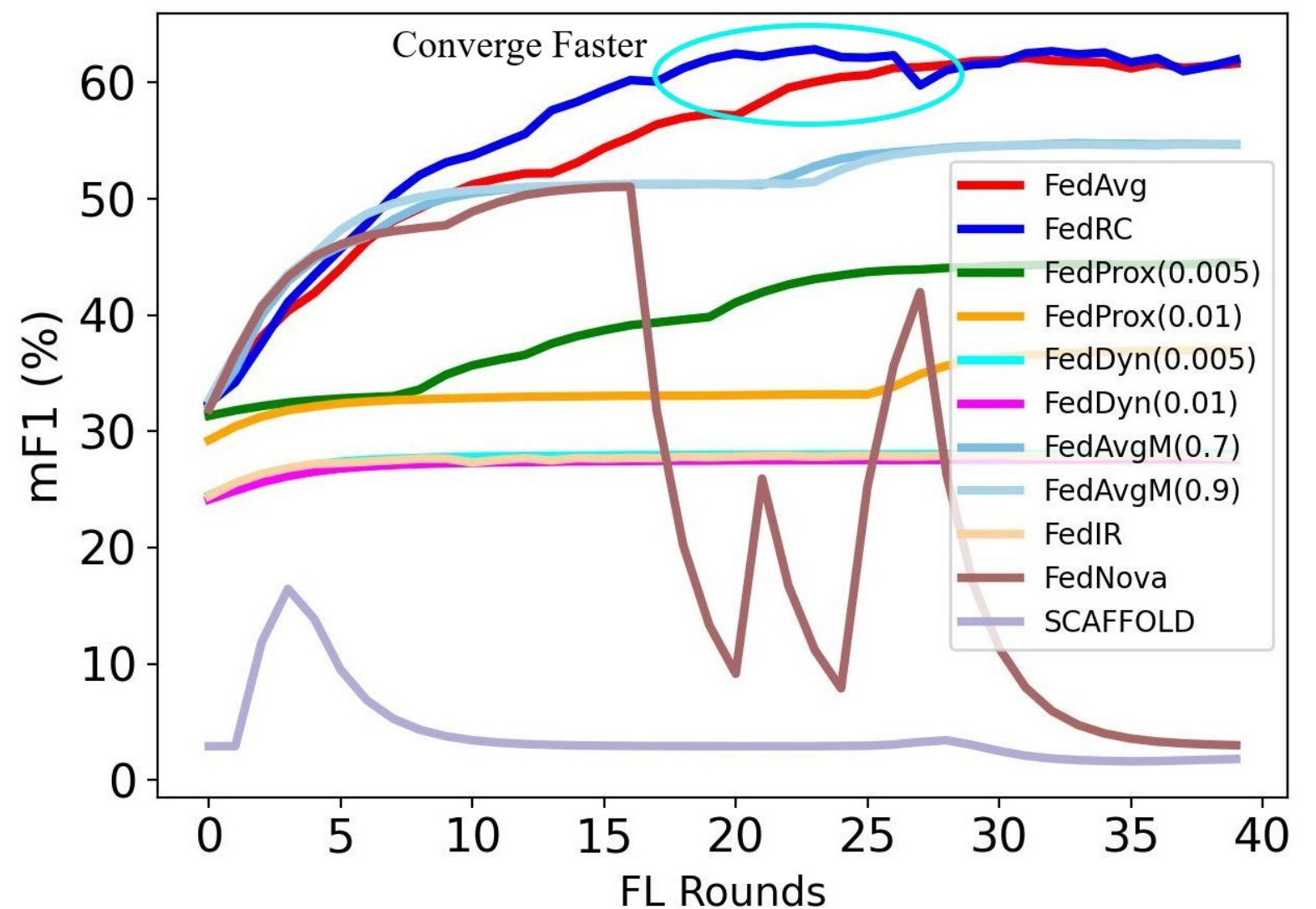}
\label{Fig.Metrics_mF1}
}
\vspace{-0.2cm}
\caption{Convergence comparison. Results show that FedRC converges faster than all other FL algorithms across all metrics.}
\label{Fig.Metrics}
\vspace{-0.5cm}
\end{figure*}

\begin{table*}[tp]
\centering
\setlength{\tabcolsep}{12.0pt}
\caption{Metrics on both Cityscapes and CamVid dataset driven by \textbf{DeepLabv3+} model}
\begin{tabularx}{\linewidth}{ccccccccc}
\hline
\multirow{2}{*}{FL Algorithms}                                   & \multicolumn{4}{c}{Cityscapes Dataset (19 Semantic Classes) (\%)}                                                                                          & \multicolumn{4}{c}{CamVid Dataset (11 Semantic Classes) (\%)}              \\ \cline{2-9} 
                                                                 & mIoU                               & mF1                                & mPrecision                         & mRecall                            & mIoU           & mF1            & mPrecision     & mRecall        \\ \hline
FedAvg \cite{https://doi.org/10.48550/arxiv.1602.05629}        & 53.61       & 62.49       & 68.90       & 59.06       & 76.72       & 85.59       & 89.89       & 84.45    \\
FedProx (0.005) \cite{li2020federated}                   & 41.51       & 47.22       & 50.22       & 46.78       & 75.46       & 82.10       & 82.46       & 81.78    \\
FedProx (0.01) \cite{li2020federated}                    & 33.67       & 37.24       & 41.86       & 38.16       & 73.57       & 80.81       & 81.47       & 80.44    \\
FedDyn (0.005) \cite{acar2021federated}               & 25.53       & 28.17       & 32.11       & 29.28       & 75.44       & 82.07       & 82.65       & 81.70    \\
FedDyn (0.01) \cite{acar2021federated}                & 24.85       & 27.64       & 26.65       & 28.77       & 64.55       & 71.60       & 80.85       & 71.55    \\
FedAvgM (0.7) \cite{hsu2019measuring}                  & 47.28       & 54.79       & 57.14       & 53.74       & 76.29       & 82.67       & 83.21       & 82.28    \\
FedAvgM (0.9) \cite{hsu2019measuring}                  & 47.17       & 54.71       & 57.07       & 53.66       & 79.23       & 87.07       & 90.03       & 85.26    \\
FedIR \cite{hsu2020federated}                                  & 25.31       & 27.94       & 27.91       & 28.46       & 60.38       & 67.27       & 77.12       & 63.89    \\
FedNova \cite{wang2020tackling}                                & 44.38       & 51.03       & 52.34       & 50.68       & 75.90       & 82.41       & 83.40       & 81.63    \\
SCAFFOLD \cite{karimireddy2020scaffold}                        & 13.55       & 16.44       & 19.76       & 17.19       & 23.74       & 30.12       & 42.85       & 31.48    \\
\textbf{\begin{tabular}[c]{@{}c@{}}FedRC (Ours)\end{tabular}} &\textbf{55.44} & \textbf{65.76} & \textbf{75.66} & \textbf{61.12} & \textbf{80.12} & \textbf{87.70} & \textbf{91.34} & \textbf{86.16} \\ \hline
\end{tabularx}
\label{Tab:metrics_deeplabv3}
\vspace{-0.6cm}
\end{table*}

\subsection{Main Results and Empirical Analysis}
\subsubsection{Convergence comparison}
In our research, we evaluate the convergence rate of the proposed FedRC algorithm against other FL algorithms based on Cityscapes and CamVid datasets. The curves of various metrics, as shown in \Cref{Fig.Metrics}, depict the convergence rates of all FL algorithms under consideration. From \Cref{Fig.Metrics_mIoU,Fig.Metrics_mPre,Fig.Metrics_mRec,Fig.Metrics_mF1}, it is obvious that FedAvg, FedRC, and both configurations of FedAvgM (FedAvgM(0.7) and FedAvgM(0.9)) outperform the rest of benchmarks with significant margins. Therefore, the following comparisons will focus on these four FL strategies. At the onset of training, FedAvgM(0.7) and FedAvgM(0.9) exhibit a steeper initial increase for all metrics compared to FedAvg and FedRC. However, as training progresses, the increasing speed of FedAvg and FedRC surpasses that of FedAvgM(0.7) and FedAvgM(0.9), and this trend continues until the training ends. Overall, FedAvg and FedRC showcase a faster convergence rate compared against the other FL algorithms.

Focusing on the convergence comparison between FedRC and FedAvg as detailed in \Cref{Fig.Metrics_mIoU,Fig.Metrics_mPre,Fig.Metrics_mRec,Fig.Metrics_mF1}, FedRC consistently exhibits a faster convergence rate than that of FedAvg. To measure this, FedAvg and FedRC reach convergence at approximately the 31-th and 19-th FL rounds in \textbf{mIoU}, respectively. This indicates that FedRC's convergence rate is accelerated by (31 - 19) / 31 = 38.71\% relative to FedAvg. Similar calculations for \textbf{mPrecision}, \textbf{mRecall} and \textbf{mF1} showcase that FedRC's convergence rate is faster than that of FedAvg by 37.5\%, 35.5\%, and 40.6\%, respectively. The reason why FedRC outperforms FedAvg is that, as emphasized before, FedRC distinguishes each RGB image by analyzing them individually rather than typically treating them equally. Furthermore, it accounts for the data's volume and statistical characteristics instead of just focusing on data volume. In other word, FedAvg is a special case of FedRC when datasets on all vehicles are i.i.d. In summary, FedRC holds a substantial advantage in convergence speed over all competing FL algorithms across all metrics.

\subsubsection{Quantitative and qualitative performance comparison}
\begin{table*}[tp]
\centering
\renewcommand{\arraystretch}{0.24}
\addtolength{\tabcolsep}{-0.45pt}
\caption{Prediction performance comparison of semantic understanding driven by varieties of FL algorithms}
\begin{tabularx}{\linewidth}{|l|lllll|}
\hline
\verticaltext[23pt]{Raw RGBs} &\includegraphics[width=0.187\linewidth, height=0.10\linewidth]{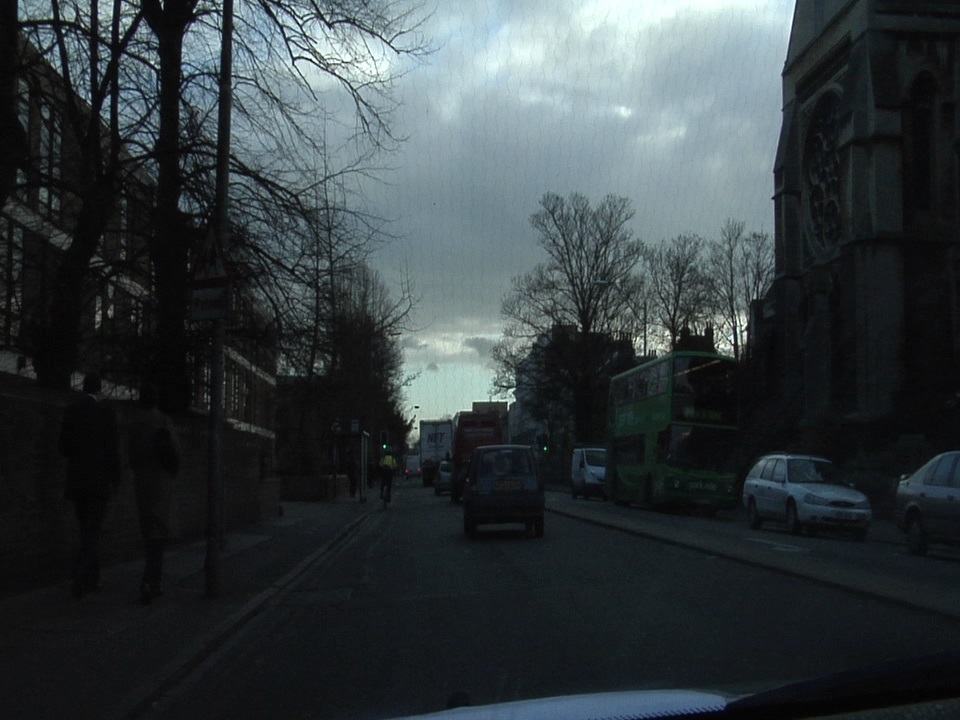} &\hspace{-0.47cm}
\includegraphics[width=0.187\linewidth, height=0.10\linewidth]{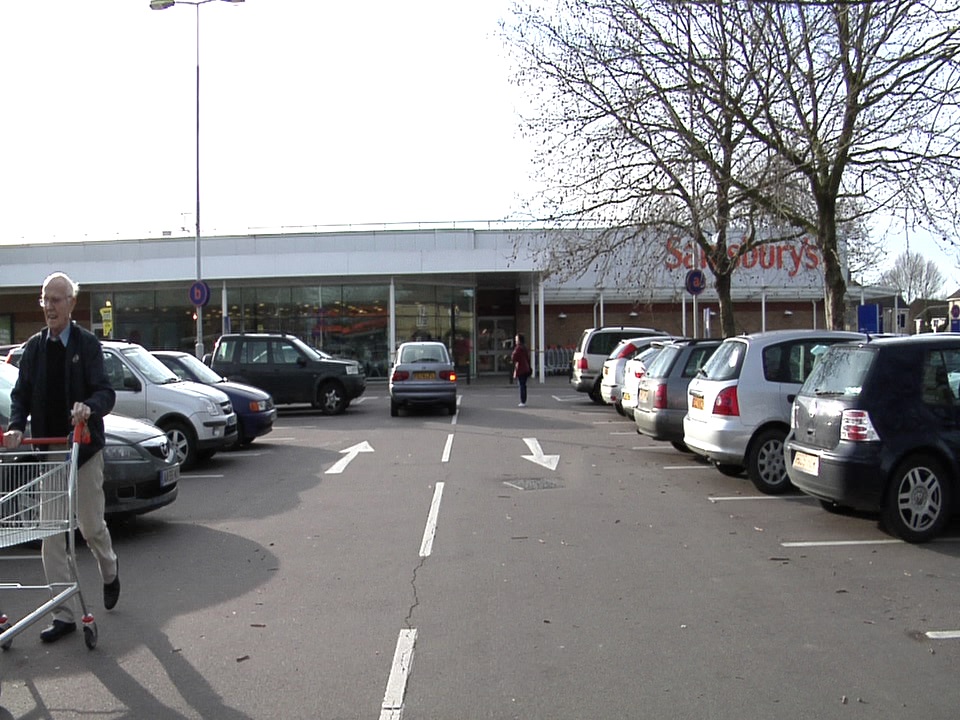} &\hspace{-0.47cm}
\includegraphics[width=0.187\linewidth, height=0.10\linewidth]{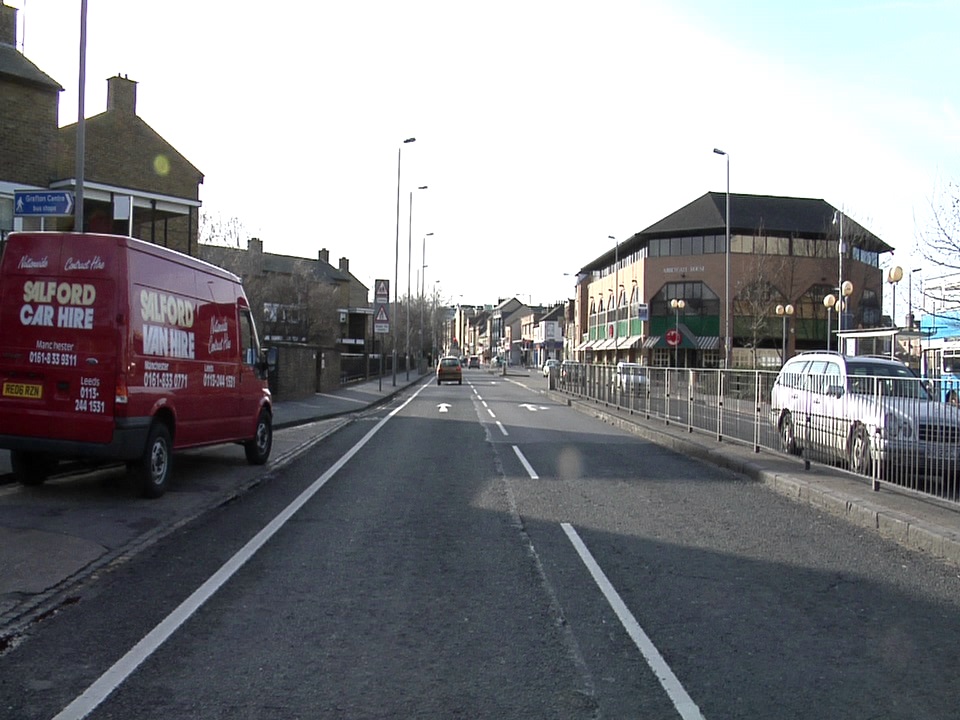} &\hspace{-0.47cm}
\includegraphics[width=0.187\linewidth, height=0.10\linewidth]{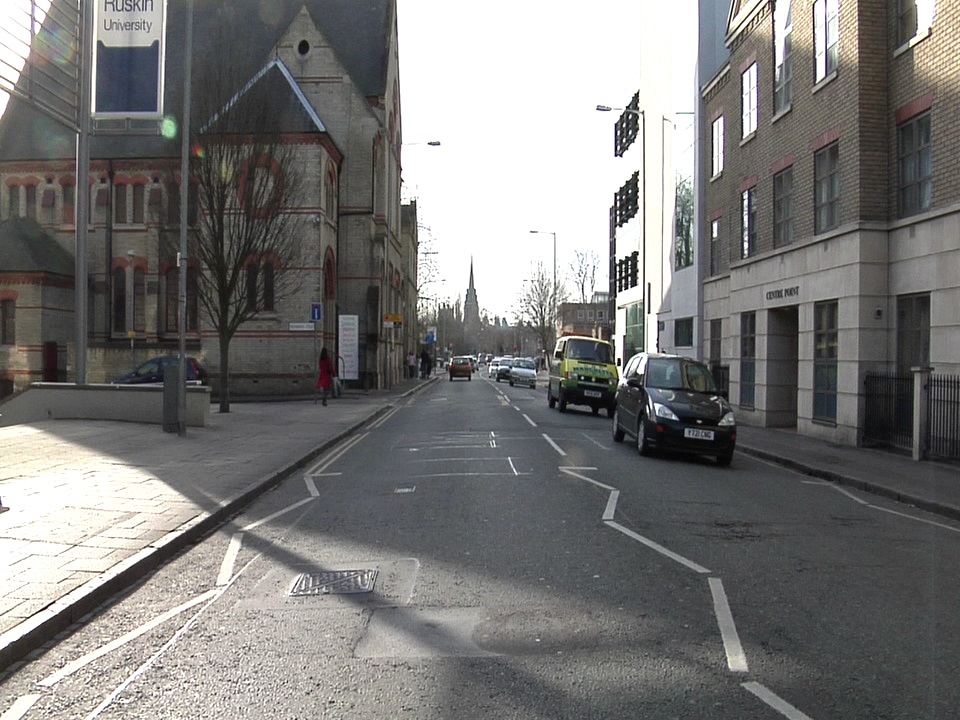} &\hspace{-0.47cm}
\includegraphics[width=0.187\linewidth, height=0.10\linewidth]{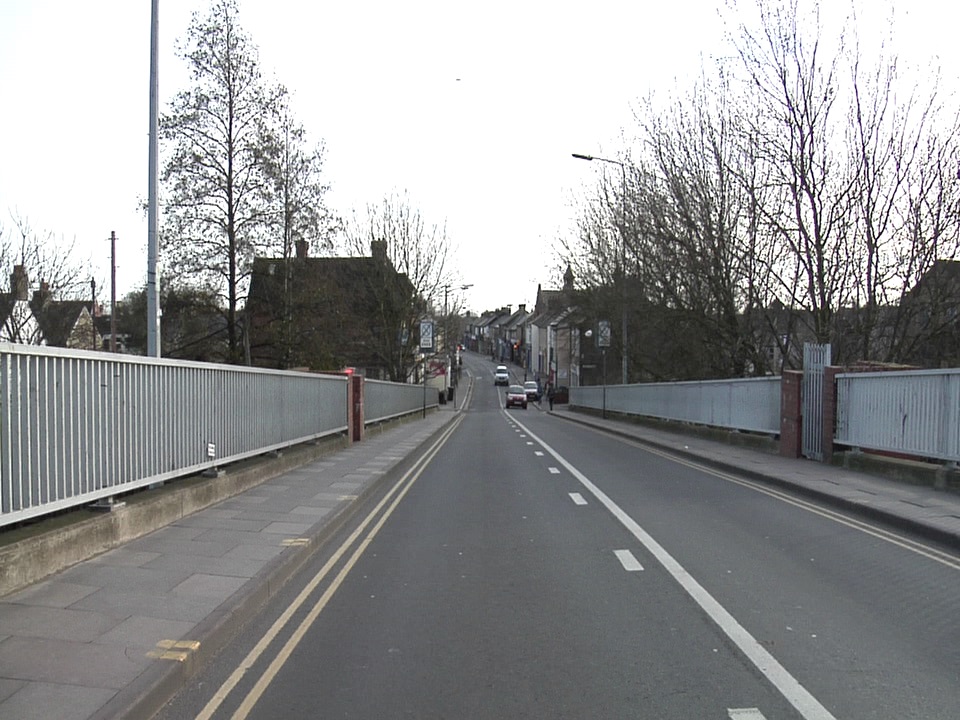}\\
\hline

\verticaltext[22pt]{Ground Truth} &\includegraphics[width=0.187\linewidth, height=0.10\linewidth]{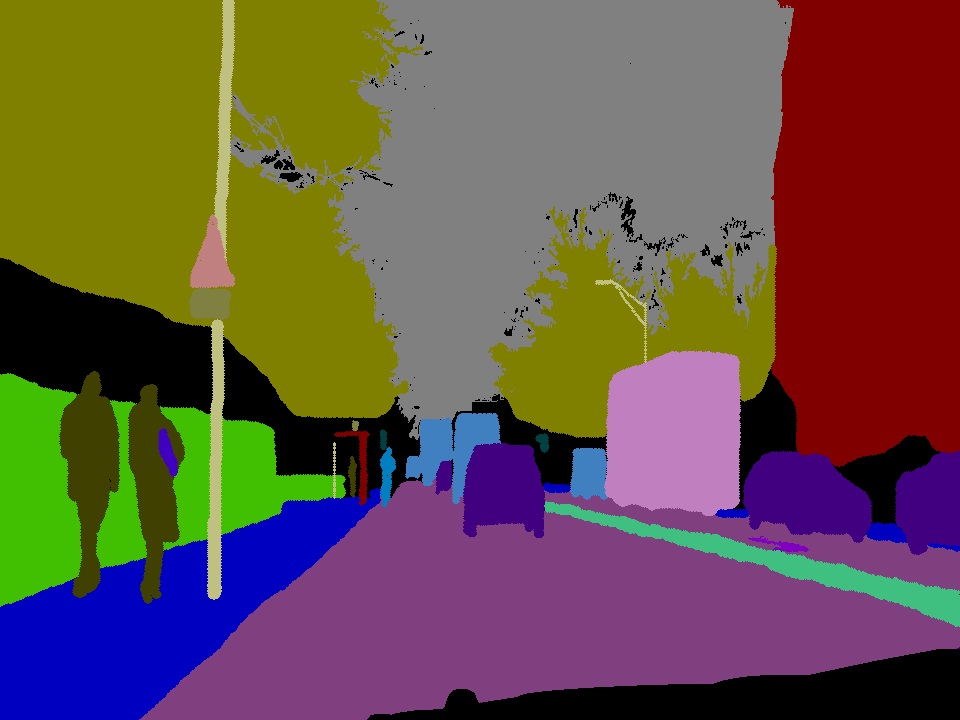} &\hspace{-0.47cm}
\includegraphics[width=0.187\linewidth, height=0.10\linewidth]{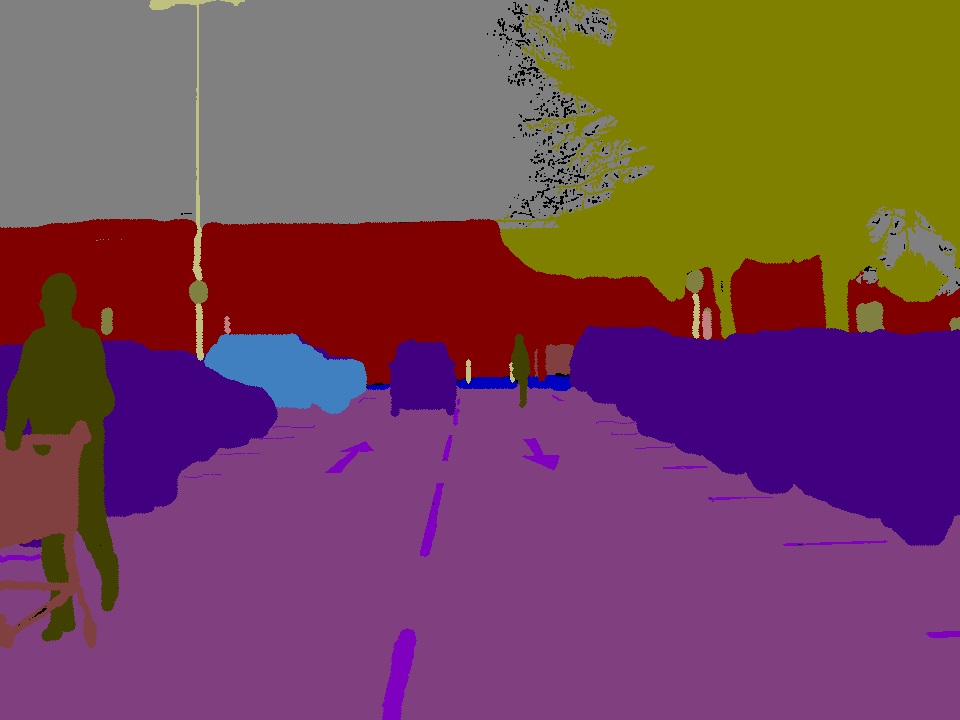} &\hspace{-0.47cm}
\includegraphics[width=0.187\linewidth, height=0.10\linewidth]{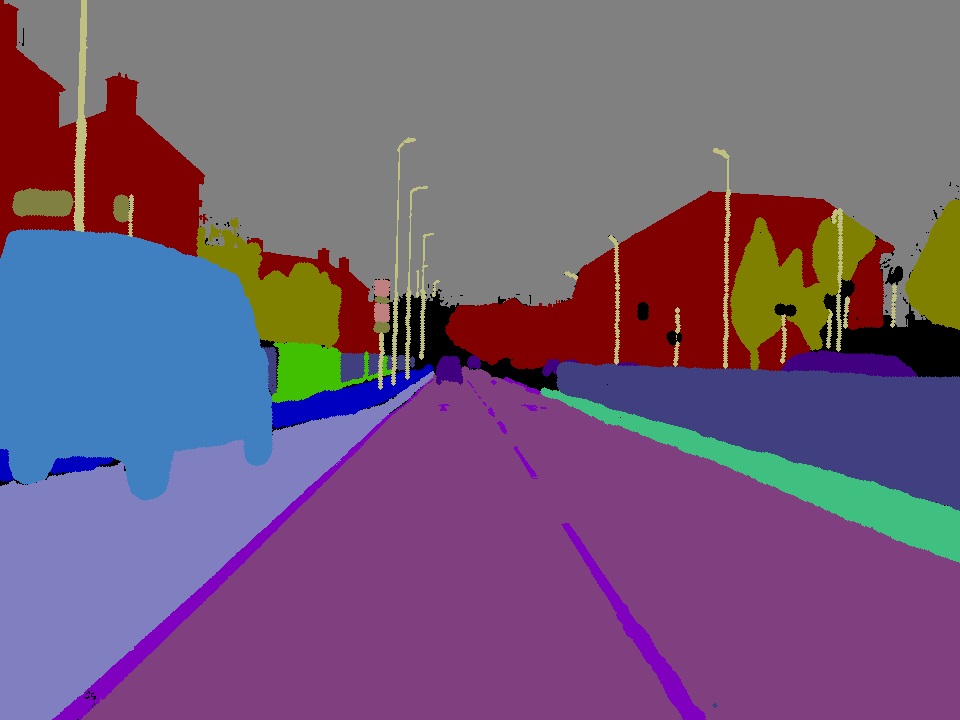} &\hspace{-0.47cm}
\includegraphics[width=0.187\linewidth, height=0.10\linewidth]{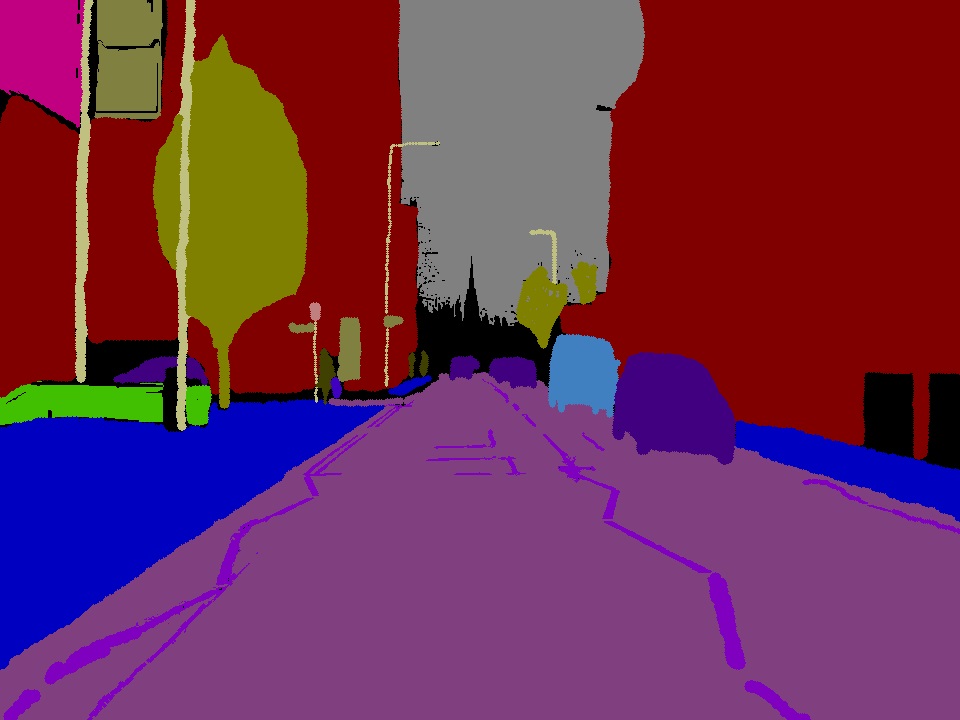} &\hspace{-0.47cm}
\includegraphics[width=0.187\linewidth, height=0.10\linewidth]{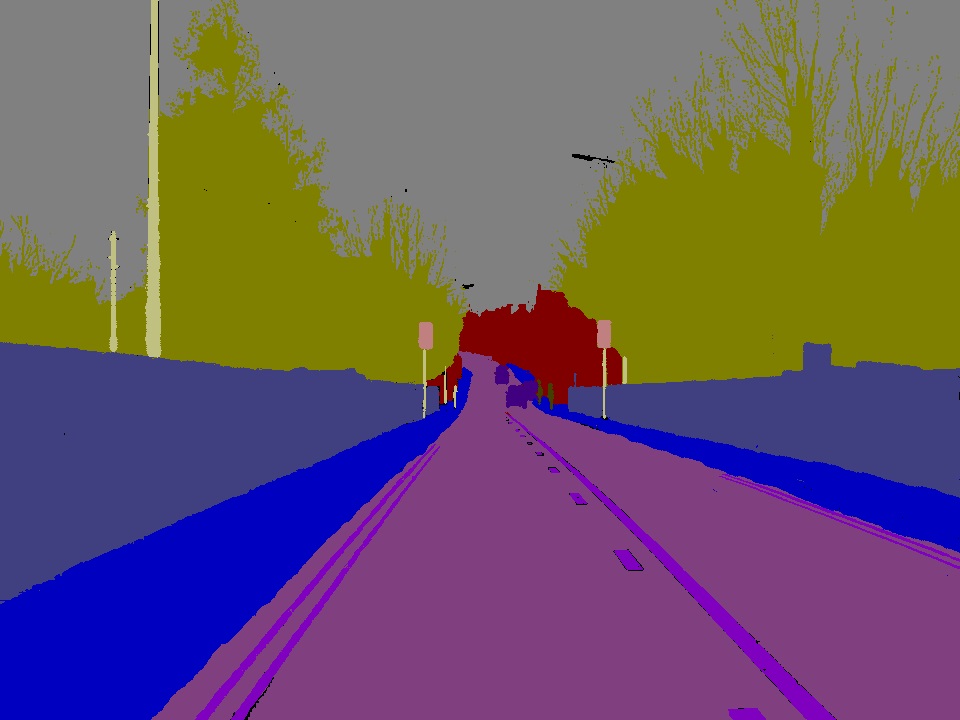}\\
\hline

\verticaltext[24pt]{FedAvg} &\includegraphics[width=0.187\linewidth, height=0.10\linewidth]{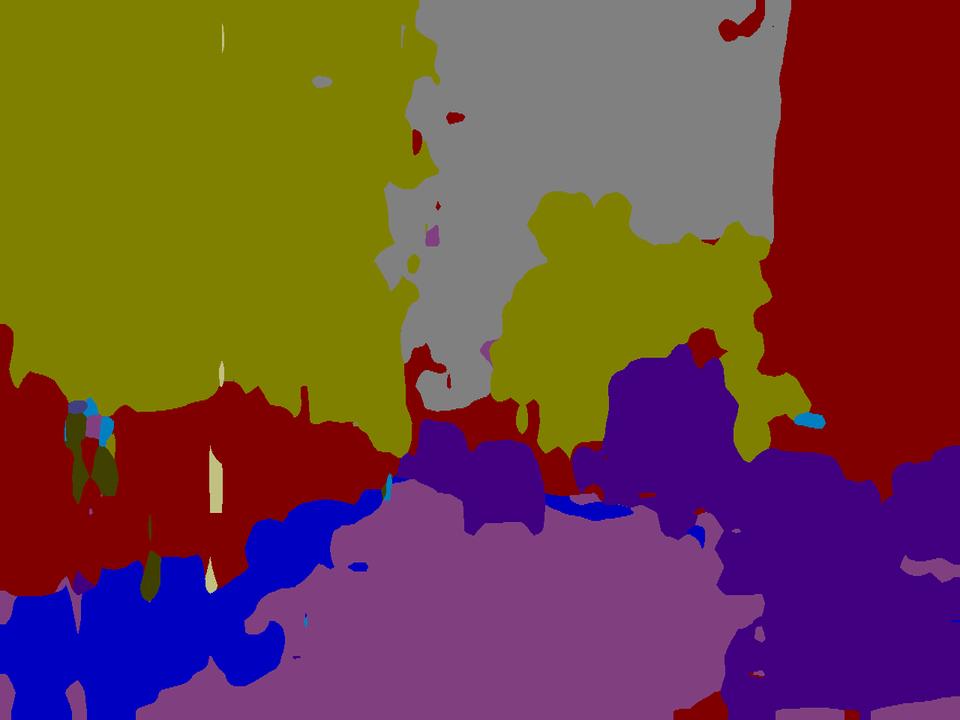} &\hspace{-0.47cm}
\includegraphics[width=0.187\linewidth, height=0.10\linewidth]{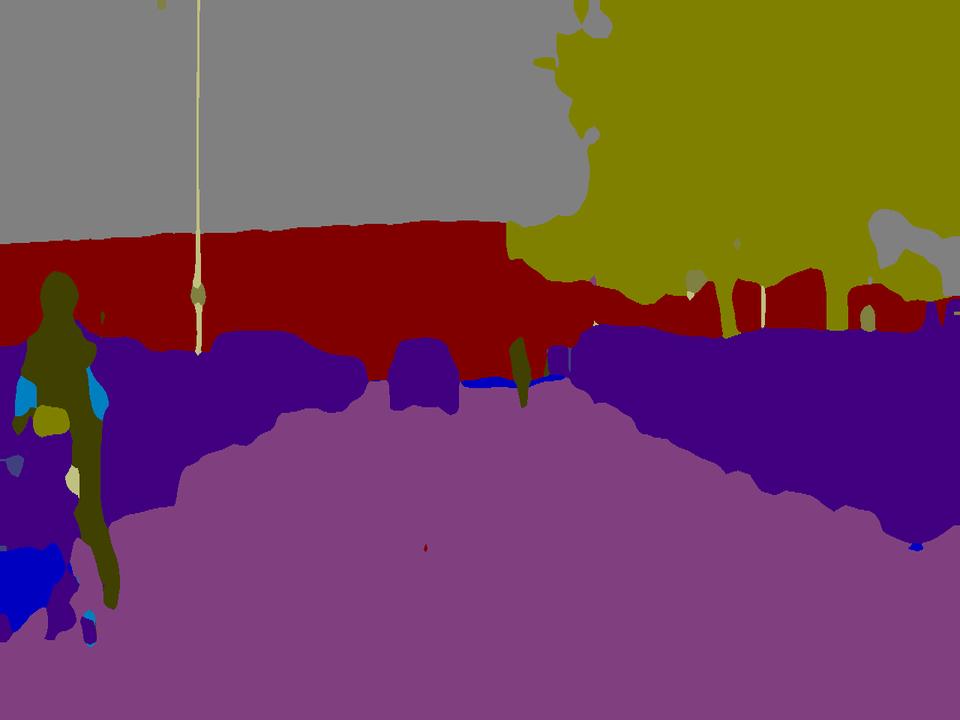} &\hspace{-0.47cm}
\includegraphics[width=0.187\linewidth, height=0.10\linewidth]{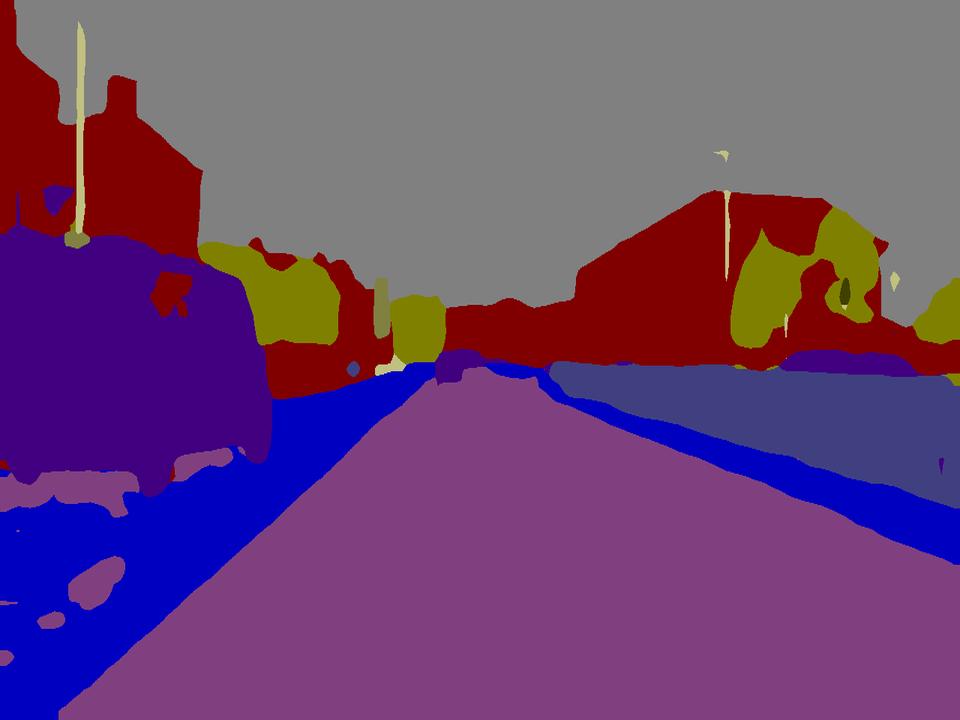} &\hspace{-0.47cm}
\includegraphics[width=0.187\linewidth, height=0.10\linewidth]{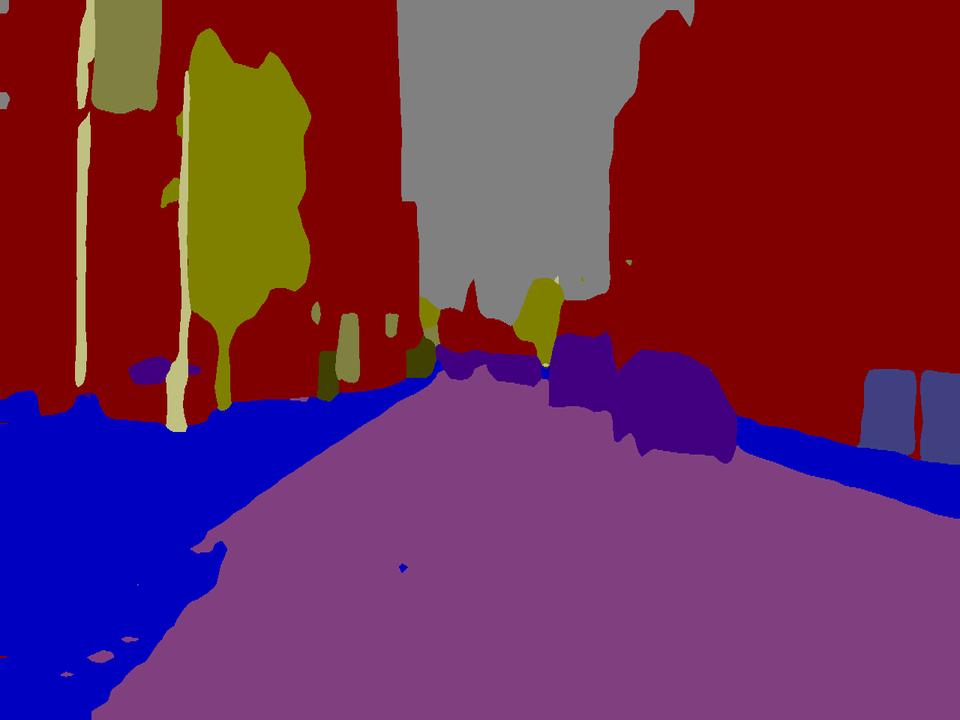} &\hspace{-0.47cm}
\includegraphics[width=0.187\linewidth, height=0.10\linewidth]{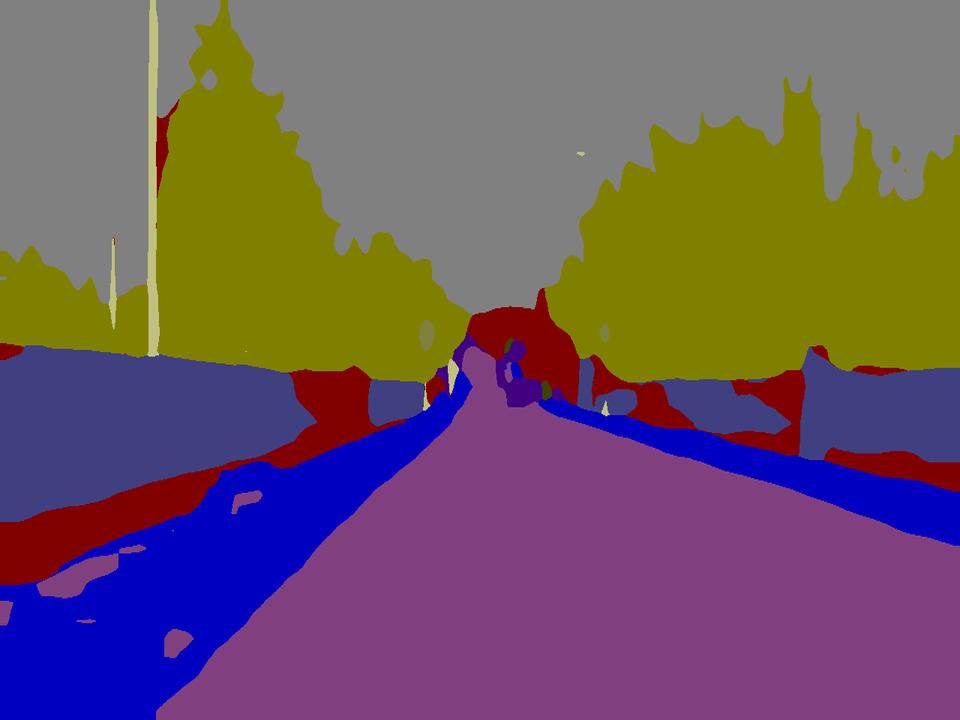}\\
\hline

\verticaltext[22.5pt]{FedAvgM(0.7)} &\includegraphics[width=0.187\linewidth, height=0.10\linewidth]{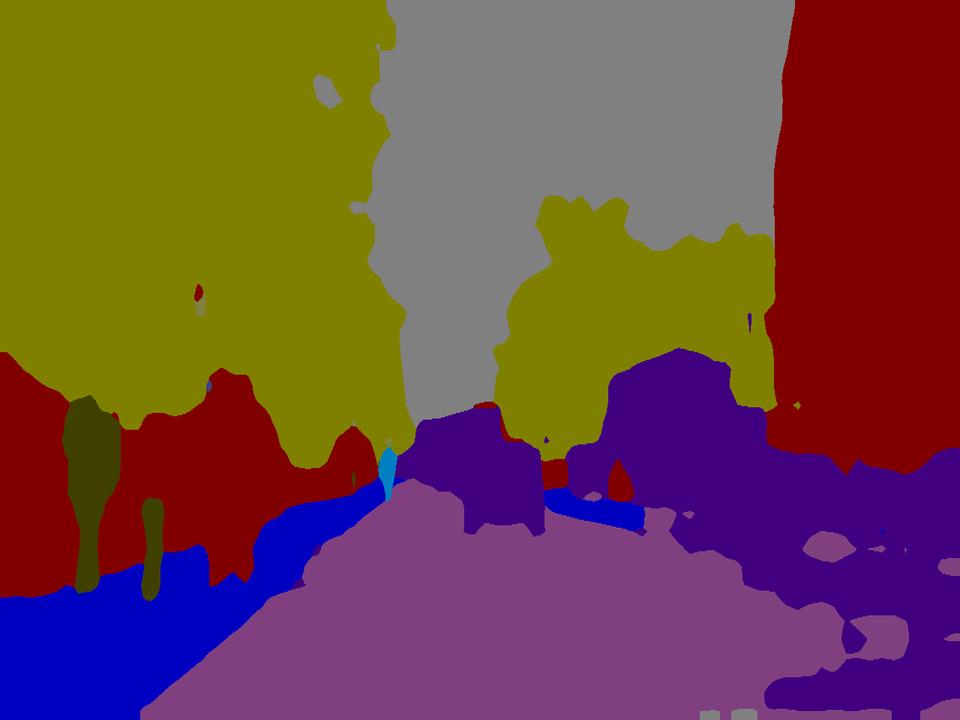} &\hspace{-0.47cm}
\includegraphics[width=0.187\linewidth, height=0.10\linewidth]{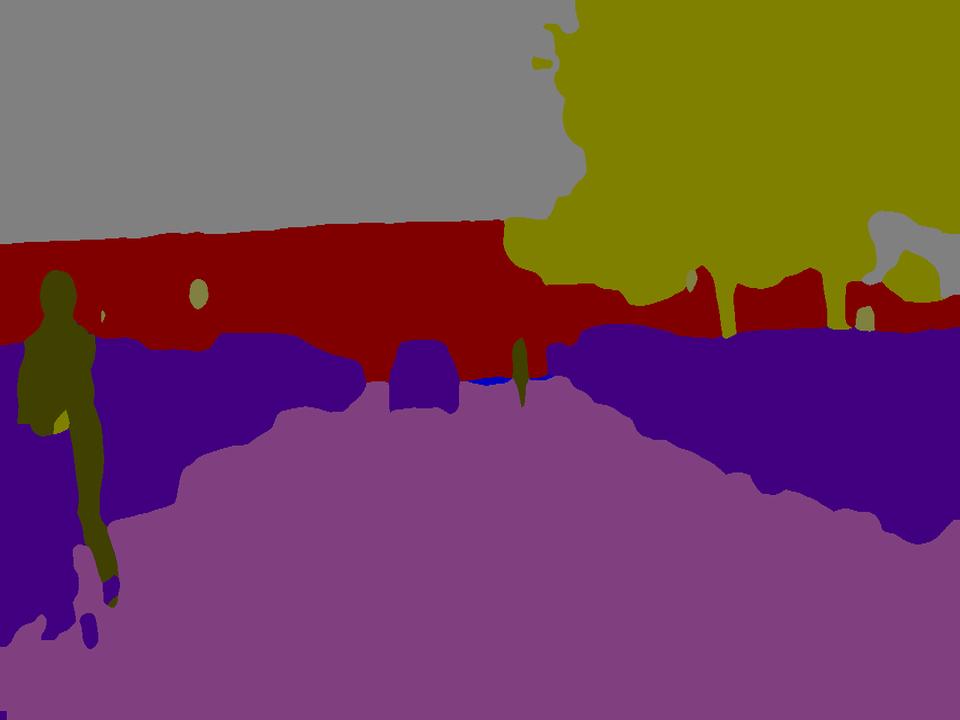} &\hspace{-0.47cm}
\includegraphics[width=0.187\linewidth, height=0.10\linewidth]{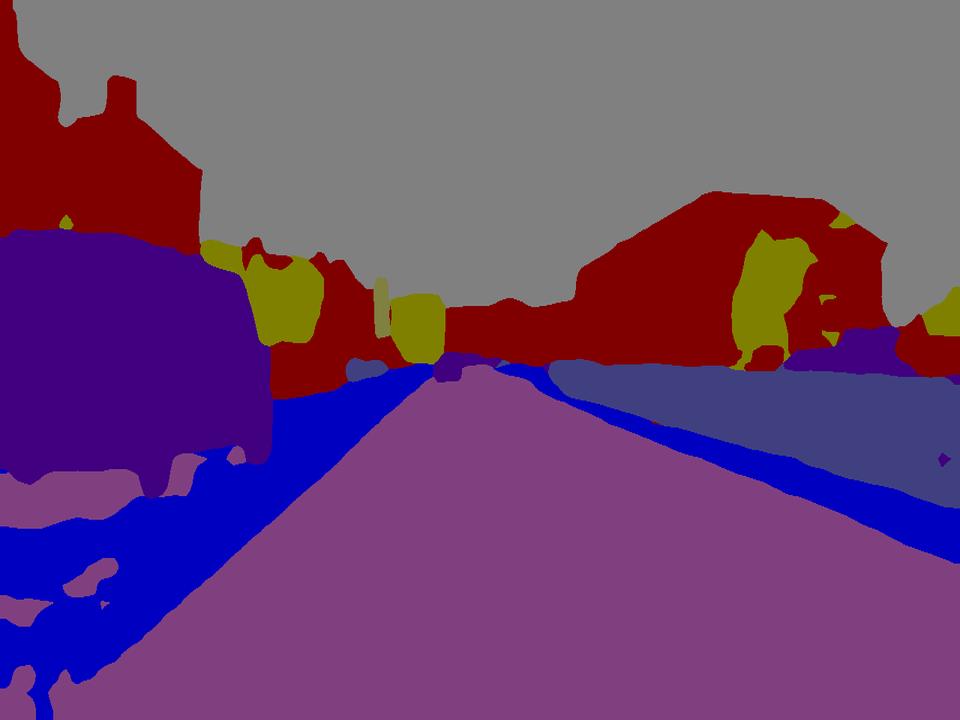} &\hspace{-0.47cm}
\includegraphics[width=0.187\linewidth, height=0.10\linewidth]{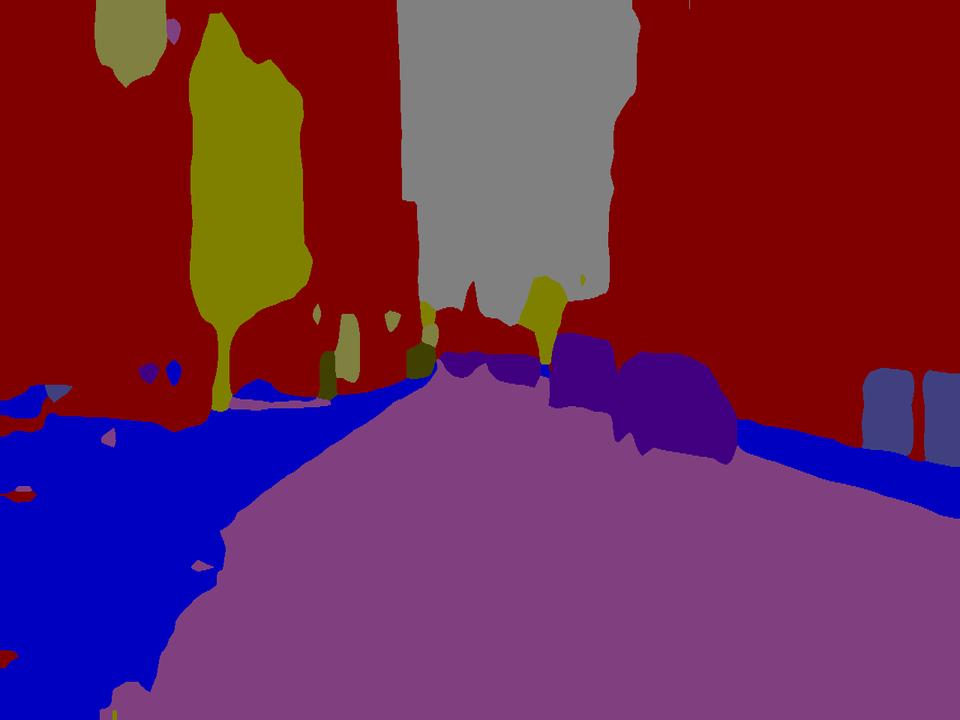} &\hspace{-0.47cm}
\includegraphics[width=0.187\linewidth, height=0.10\linewidth]{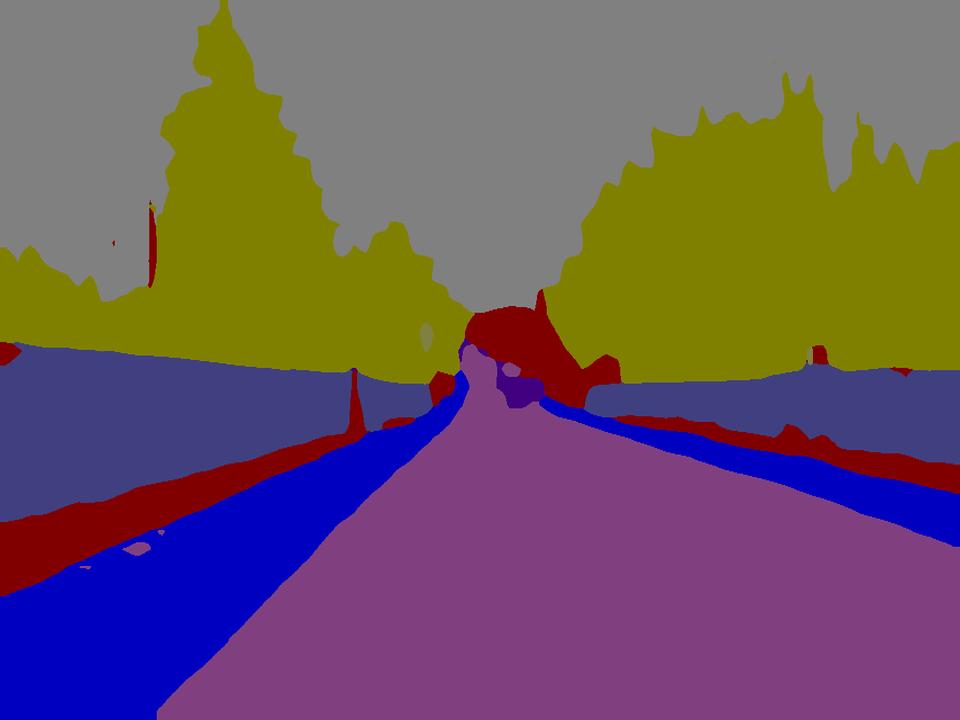}\\
\hline

\verticaltext[23pt]{FedNova} &\includegraphics[width=0.187\linewidth, height=0.10\linewidth]{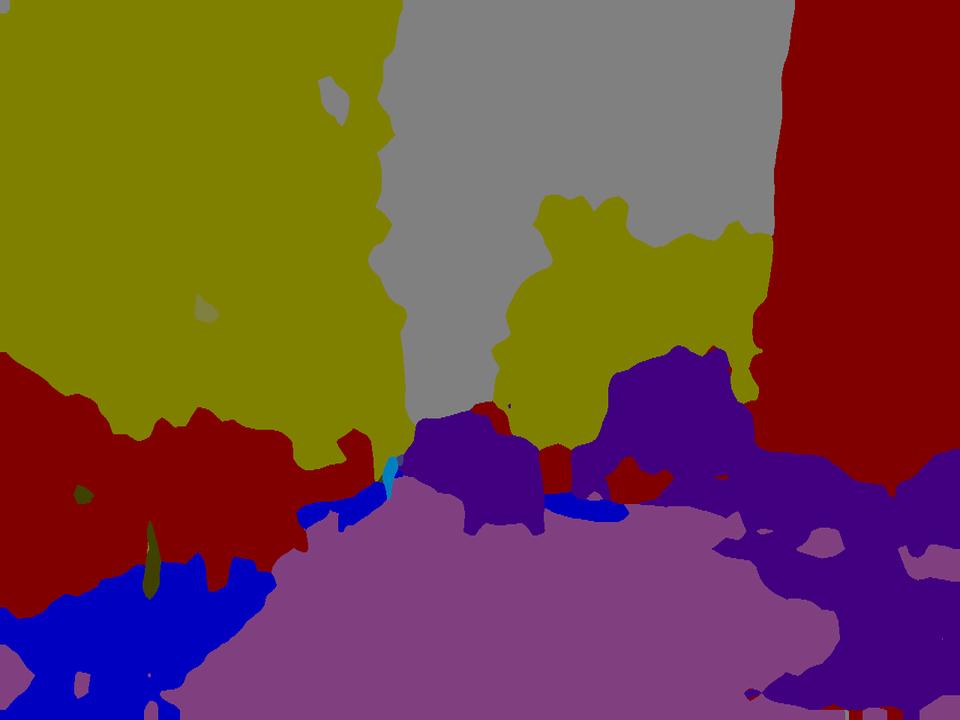} &\hspace{-0.47cm}
\includegraphics[width=0.187\linewidth, height=0.10\linewidth]{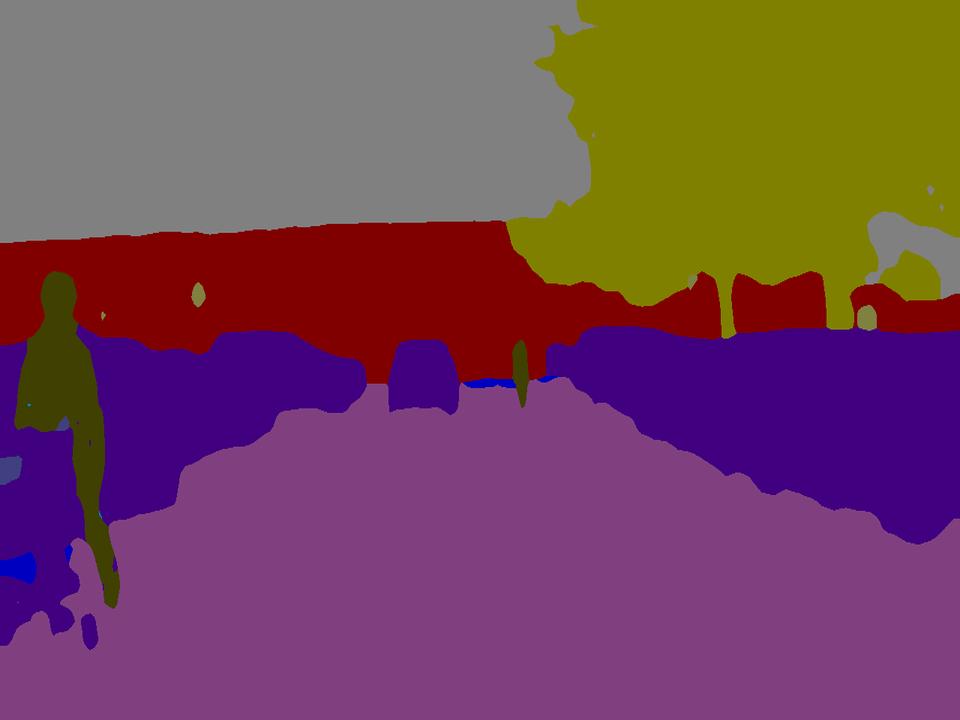} &\hspace{-0.47cm}
\includegraphics[width=0.187\linewidth, height=0.10\linewidth]{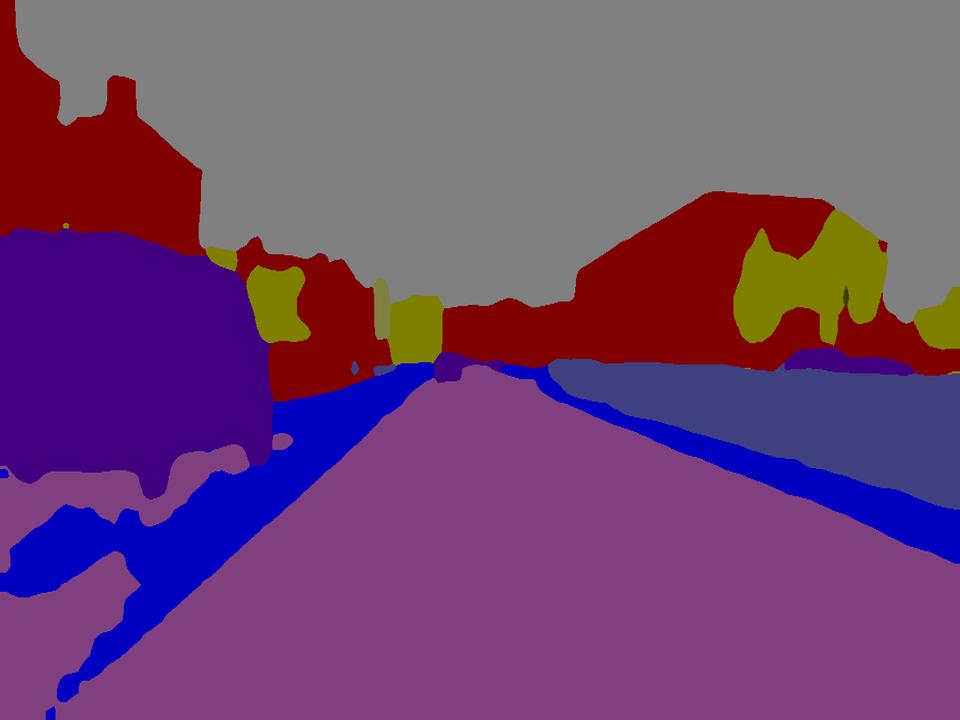} &\hspace{-0.47cm}
\includegraphics[width=0.187\linewidth, height=0.10\linewidth]{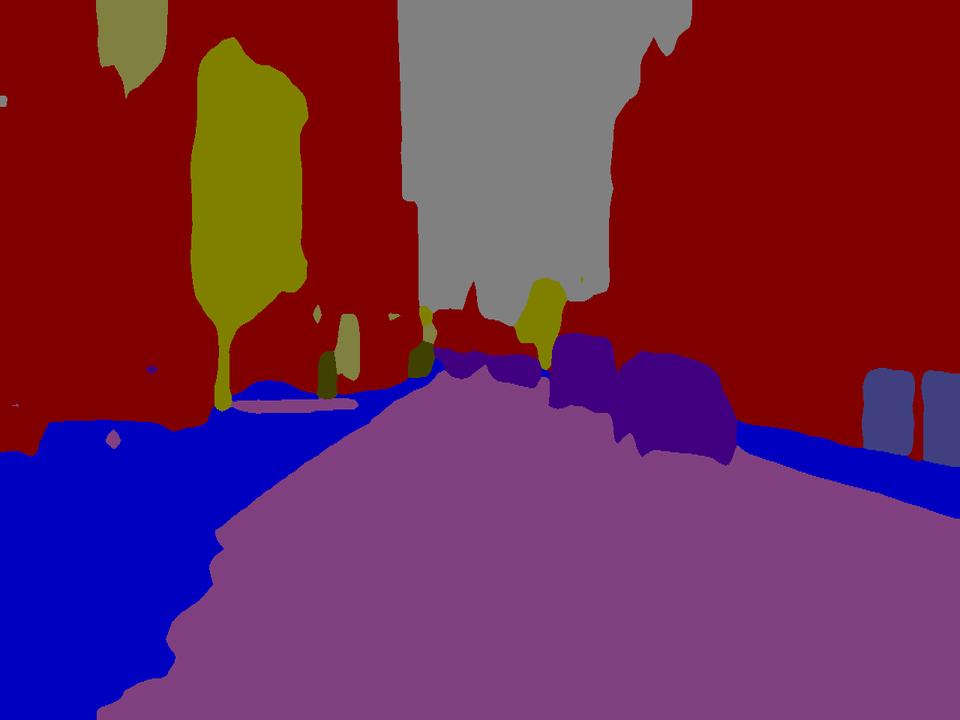} &\hspace{-0.47cm}
\includegraphics[width=0.187\linewidth, height=0.10\linewidth]{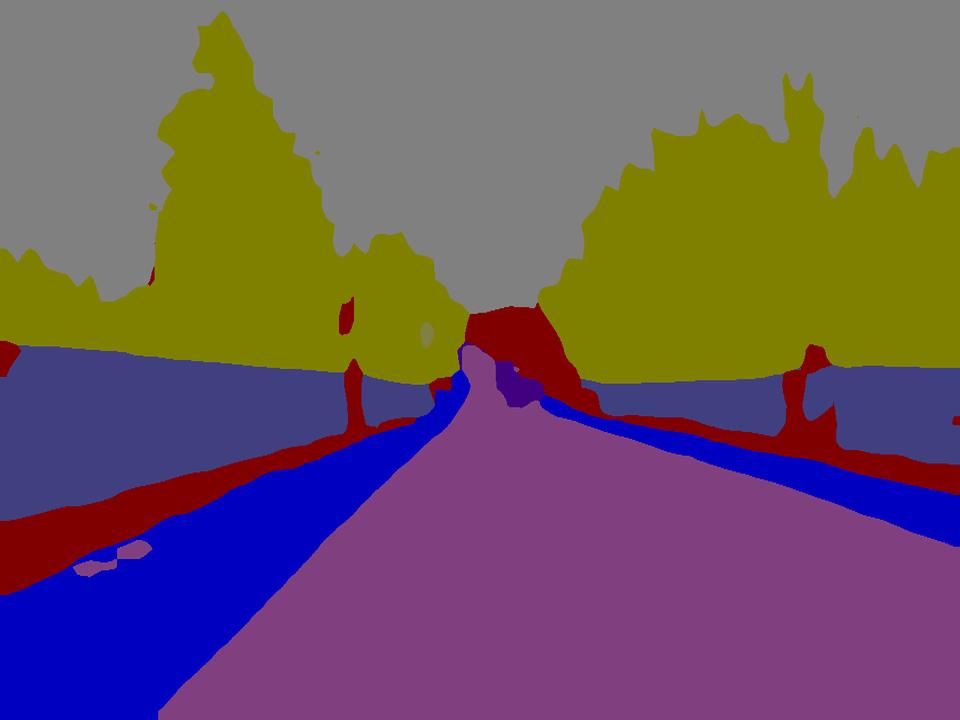}\\
\hline

\verticaltext[22.5pt]{\textbf{FedRC (ours)}} &\includegraphics[width=0.187\linewidth, height=0.10\linewidth]{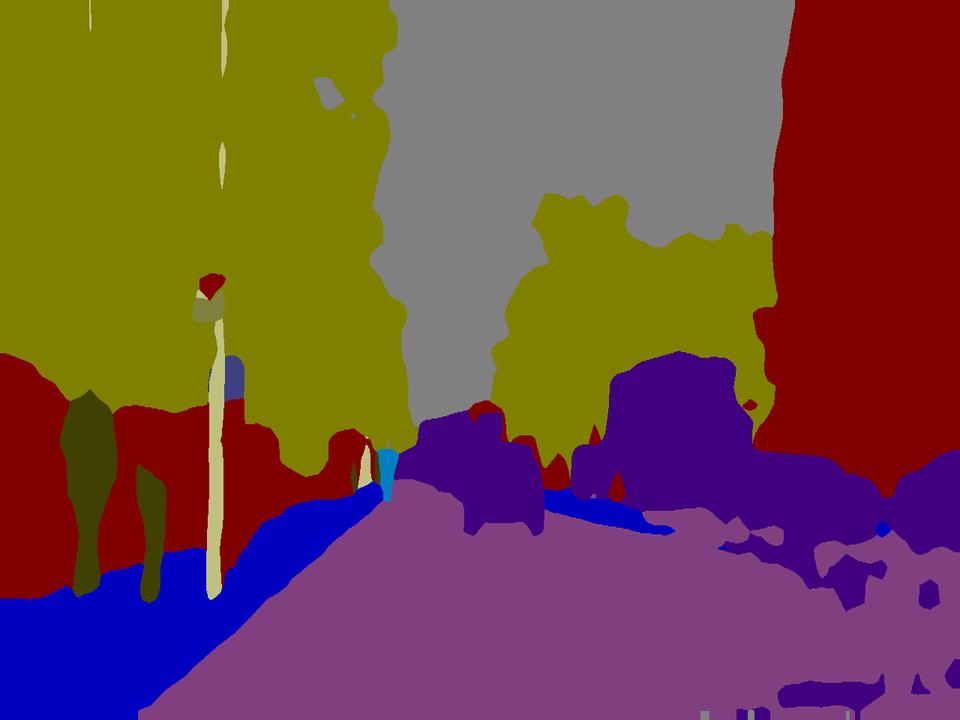} &\hspace{-0.47cm}
\includegraphics[width=0.187\linewidth, height=0.10\linewidth]{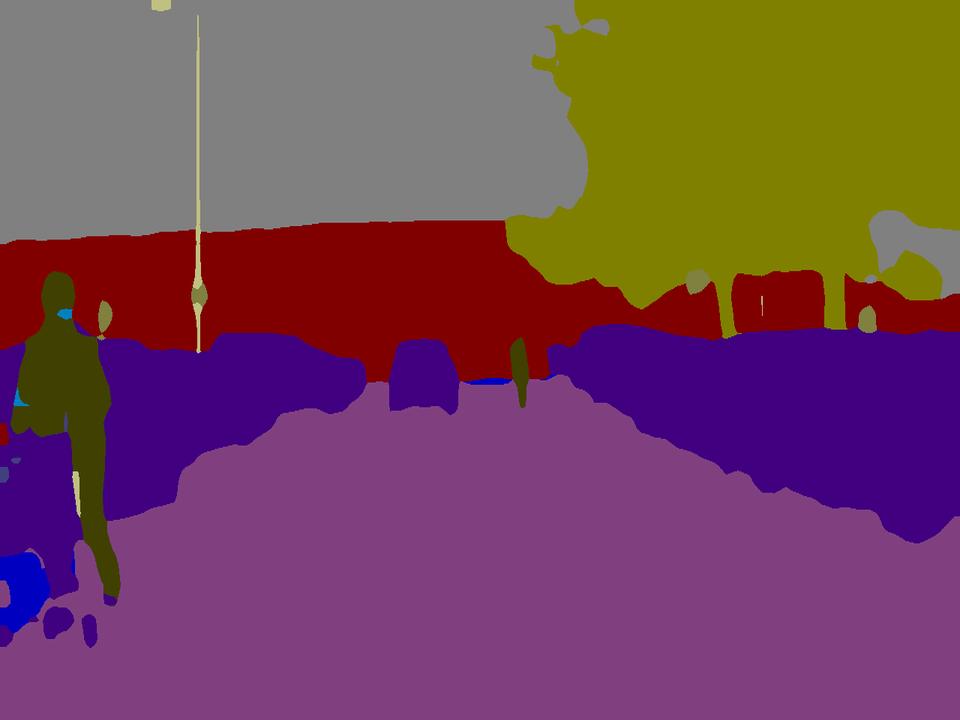} &\hspace{-0.47cm}
\includegraphics[width=0.187\linewidth, height=0.10\linewidth]{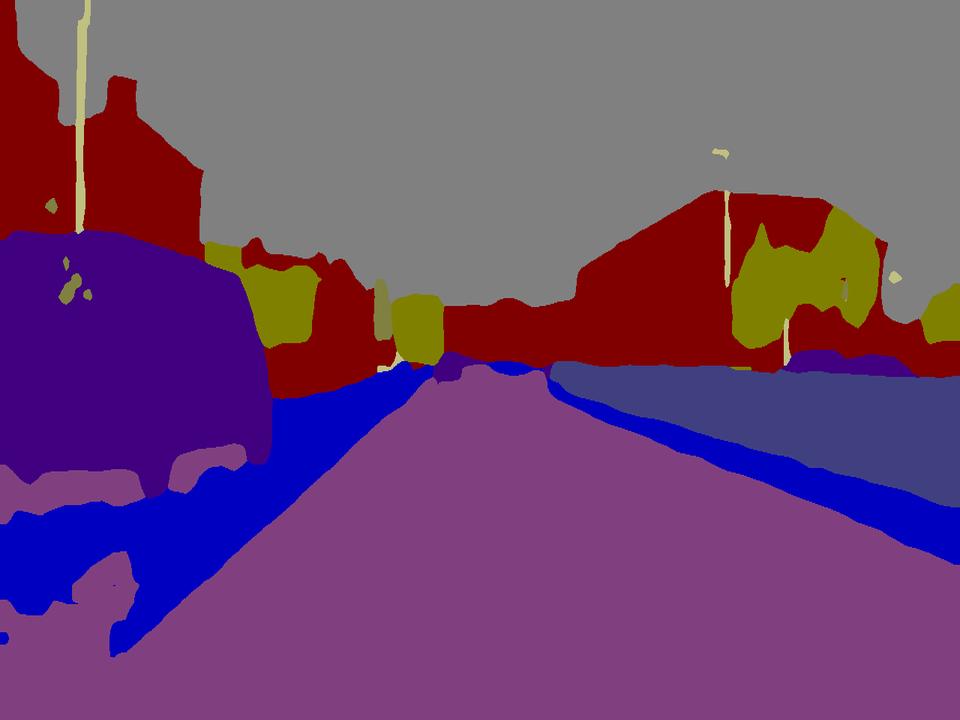} &\hspace{-0.47cm}
\includegraphics[width=0.187\linewidth, height=0.10\linewidth]{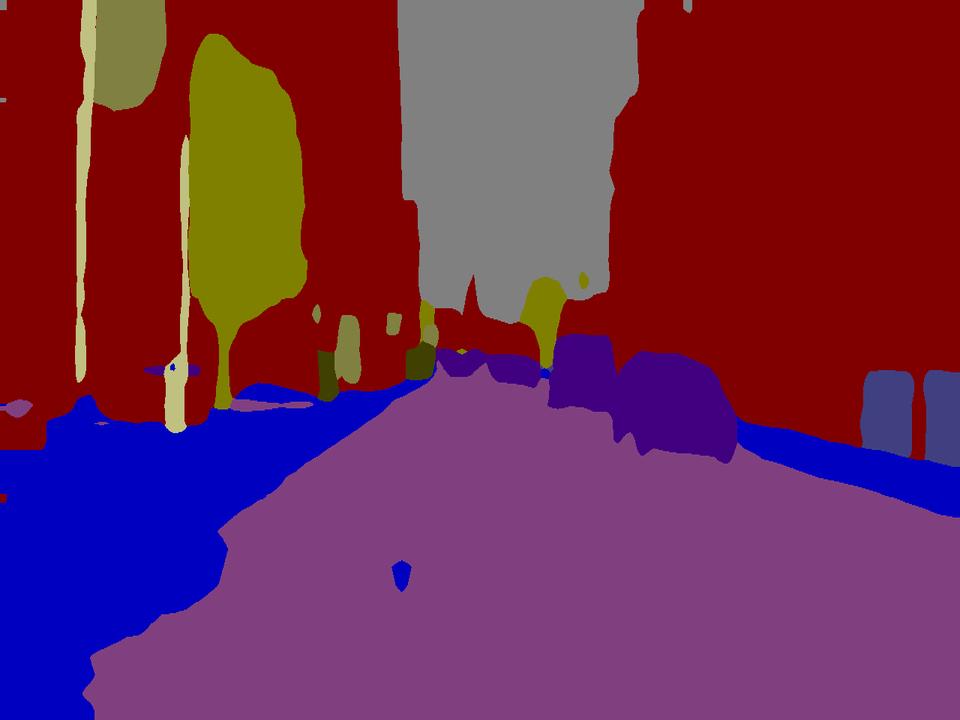} &\hspace{-0.47cm}
\includegraphics[width=0.187\linewidth, height=0.10\linewidth]{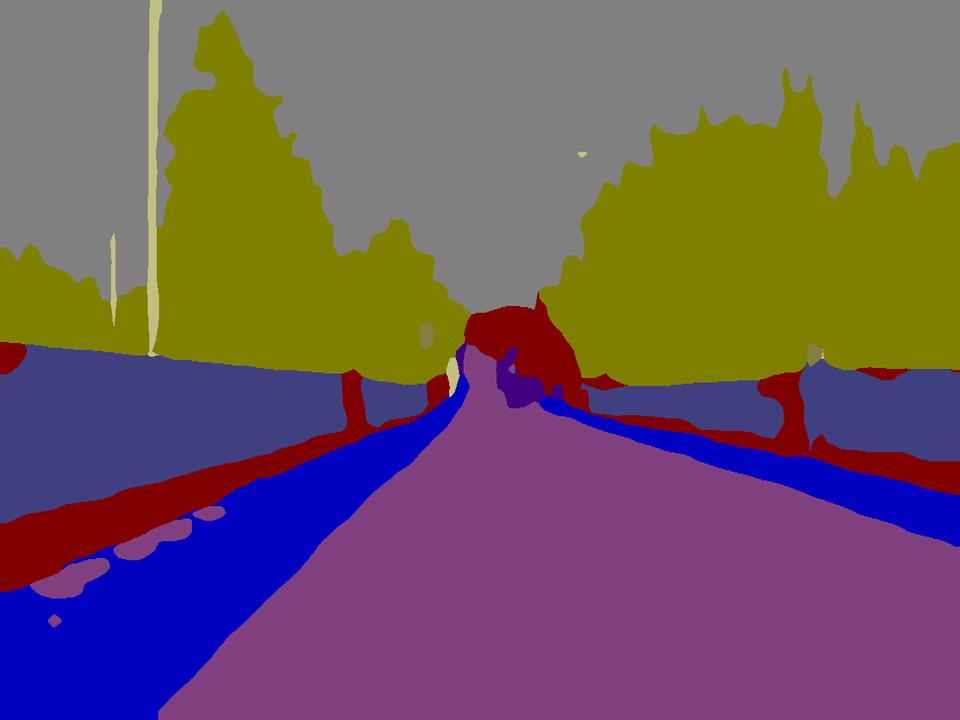}\\ \hline
\end{tabularx}
\label{tab:semantic_pred}
\vspace{-0.6cm}
\end{table*}

In the \textbf{Quantitatively} analysis, we carry out a set of experiments to benchmark the performance of various FL algorithms driven by DeepLabv3+ model \cite{chen2018encoderdecoder}. The results for the DeepLabv3+ model are presented in \Cref{Tab:metrics_deeplabv3}, which clearly indicates that FedRC exceeds all other algorithms in performance across almost all metrics for both Cityscapes and CamVid datasets. Specifically, for Cityscapes dataset, FedRC outperforms the second-best FL algorithm (\ie, FedAvg) by margins of (55.44 - 53.61) \%  = 1.83\%, (65.76 - 62.49) \%  = 3.27\%, (75.66 - 68.90) \%  = 6.76\% and (61.12 - 59.06) \%  = 2.06\% in mIoU, mPrecision, mRecall, and mF1, respectively. For CamVid dataset, the improvements of FedRC over FedAvg are (80.12 - 76.72) \%  = 3.40\%, (87.70 - 85.59) \%  = 2.11\%, (91.34 - 89.89) \%  = 1.45\% and (86.16 - 84.45) \%  = 1.71\% across mIoU, mPrecision, mRecall and mF1. 
On the other hand, upon inspecting \Cref{Tab:metrics_deeplabv3}, it suggests that a negative correlation between the performance of FL algorithms and task complexity. Algorithms like FedProx, FedDyn, and FedNova, for example, show superior outcomes on relatively easy classification task, yet lag behind on more complicated TriSU task. This pattern of inverse correlation is also applicable when comparing the performance of FL algorithms against the complexity of the datasets utilized. For instance, the majority of FL algorithms tend to underperform on the complex Cityscapes dataset relative to their performance on the simpler CamVid dataset. 

In the \textbf{qualitative} analysis, \Cref{tab:semantic_pred} illustrates the results of various FL algorithms, including FedAvg, FedAvgM(0.7), FedDyn(0.005), FedProx(0.005), FedNova, along with our FedRC, on five RGB images from diverse AD scenarios. To measure the effectiveness of each FL algorithm's prediction performance, we examine how closely their prediction outputs align with the ground truth and the original images. The comparison reveals that FedRC's outputs are consistently more accurate in capturing both the overall scene and details for all images. For example, FedRC is the only algorithm that reliably identifies subtle elements such as poles, depicted in light yellow, which most other FL algorithms tend to overlook.

\begin{figure}[tp]
\centering 
\includegraphics[width=\linewidth, height=0.6\linewidth]{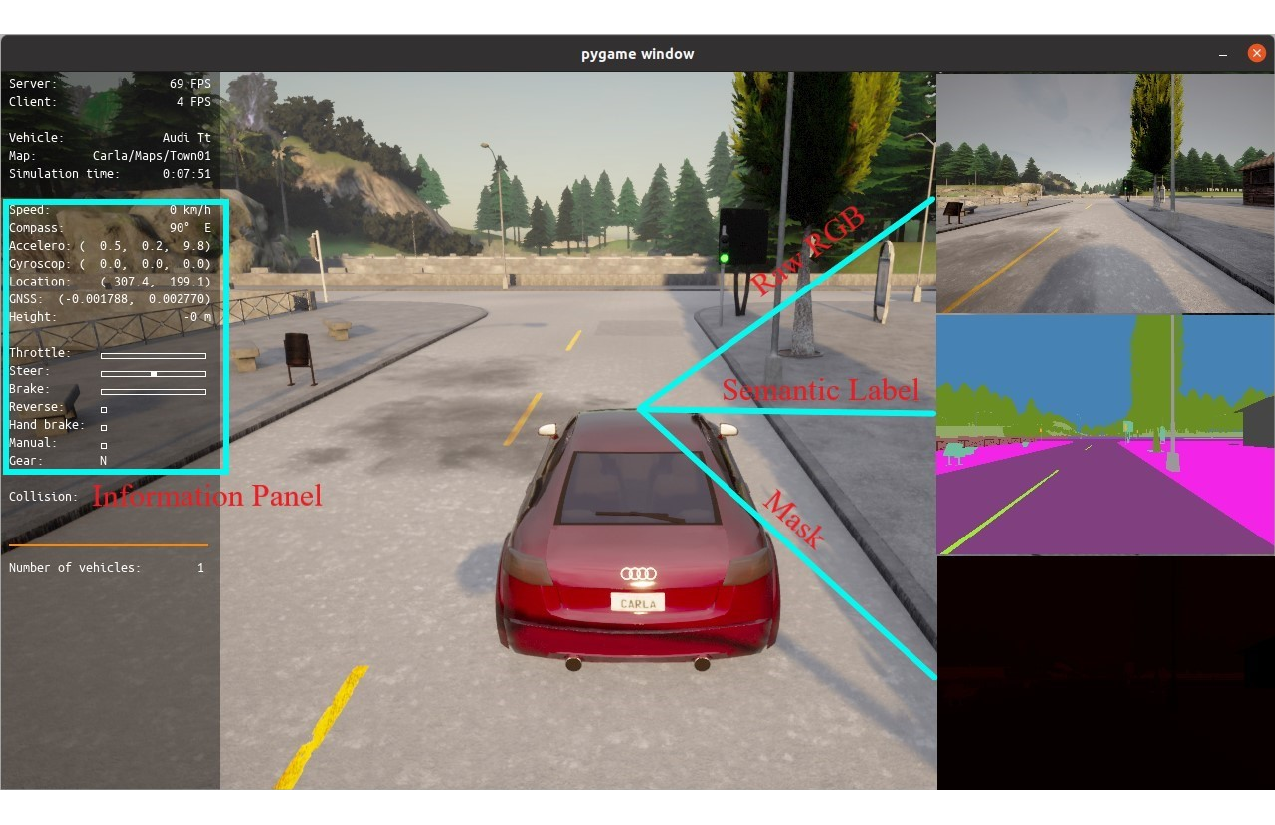}
\vspace{-0.7cm}
\caption{The demonstration of capturing CARLA data.}
\label{Fig.capturing}
\vspace{-0.2cm}
\end{figure}

\begin{table}[tp]
\centering
\renewcommand{\arraystretch}{0.24}
\addtolength{\tabcolsep}{-0.45pt}
\caption{Prediction performance comparison of varieties of models on CARLA simulation data}
\vspace{0.1cm}
\begin{tabularx}{\linewidth}{|llll|}
\hline
\hspace{0.3cm}Raw Images &\hspace{-0.15cm}Ground Truth &\hspace{0.15cm}FedAvg &\hspace{-0.15cm}FedRC (Ours) \\
\hline
\includegraphics[width=0.237\linewidth, height=0.17\linewidth]{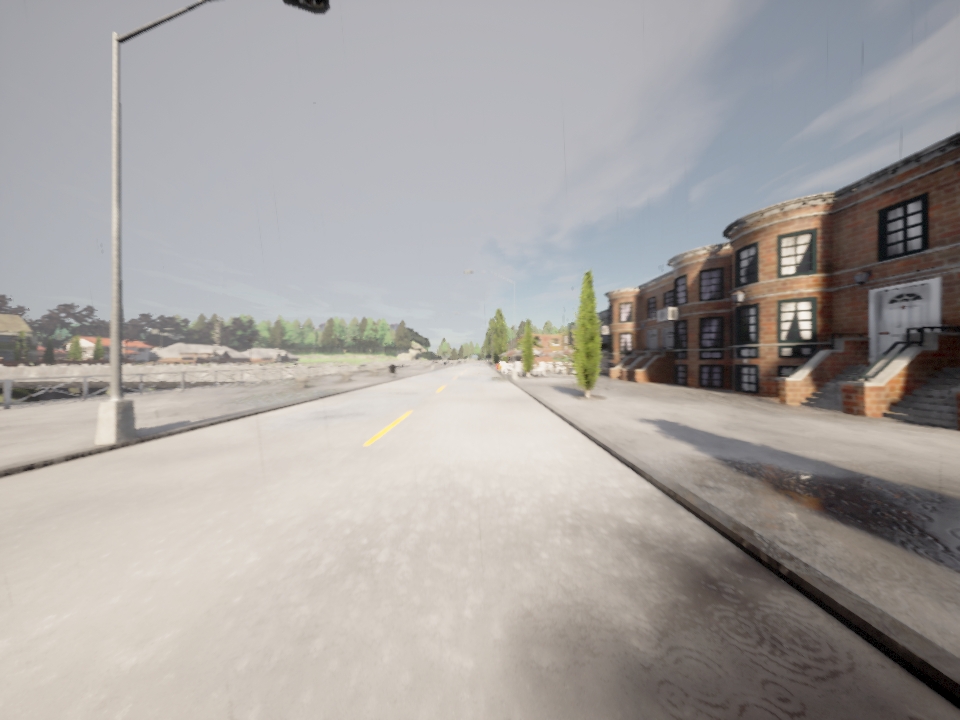} &\hspace{-0.47cm}
\includegraphics[width=0.237\linewidth, height=0.17\linewidth]{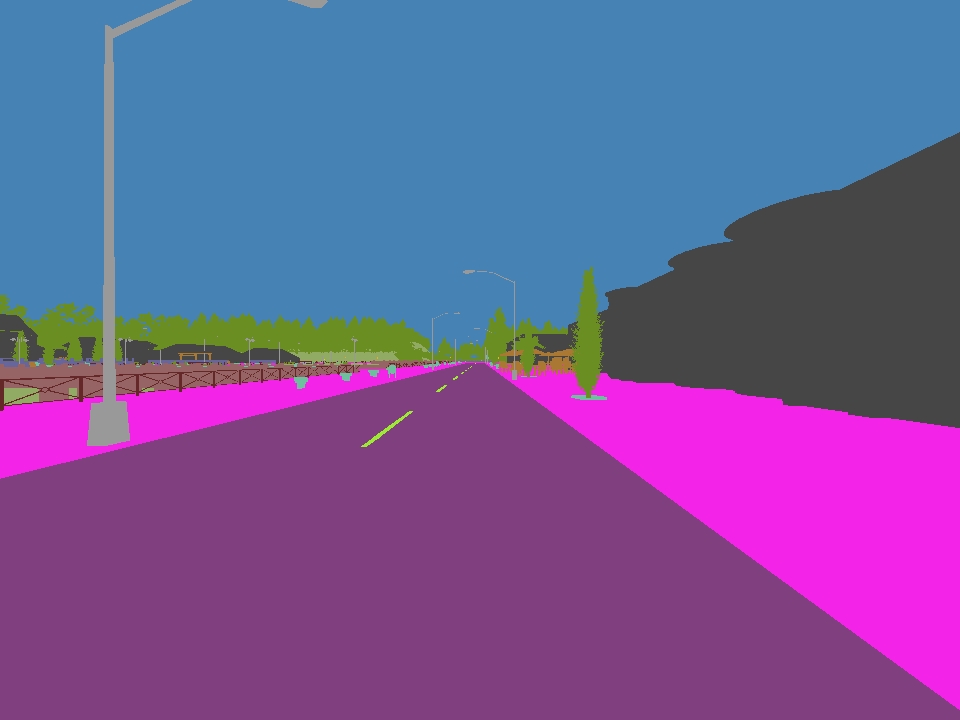} &\hspace{-0.47cm}
\includegraphics[width=0.237\linewidth, height=0.17\linewidth]{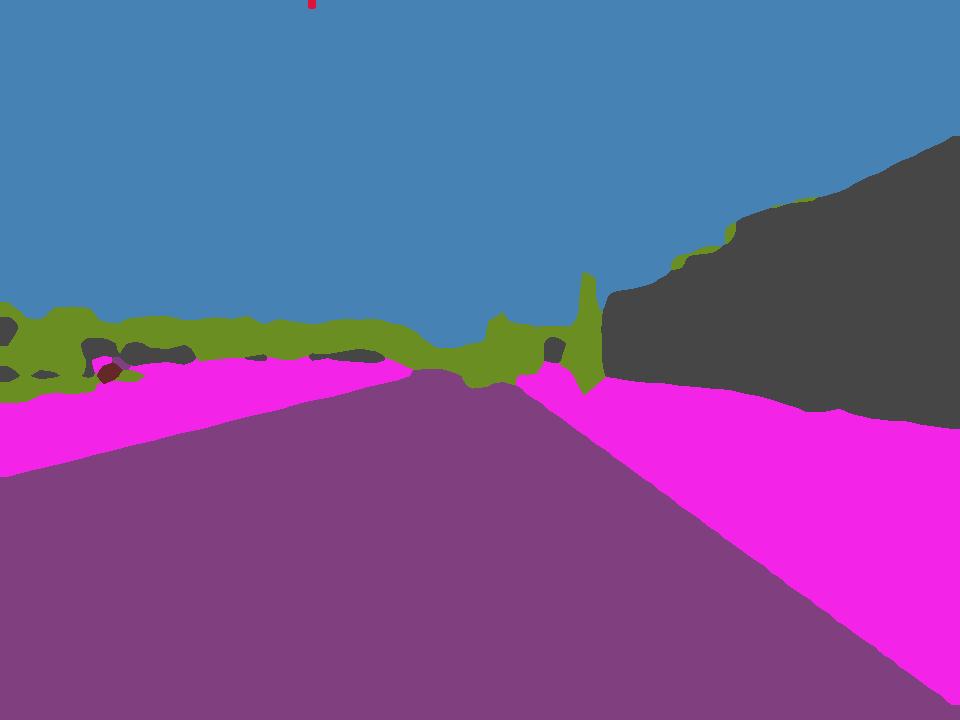} &\hspace{-0.47cm}
\includegraphics[width=0.237\linewidth, height=0.17\linewidth]{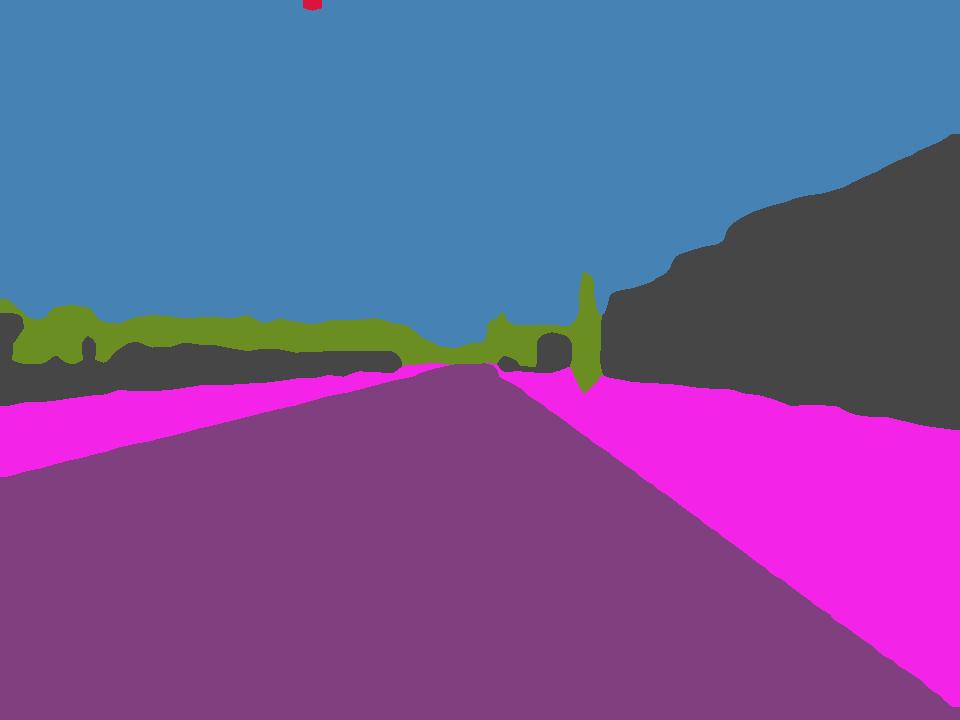}\\
\hline

\includegraphics[width=0.237\linewidth, height=0.17\linewidth]{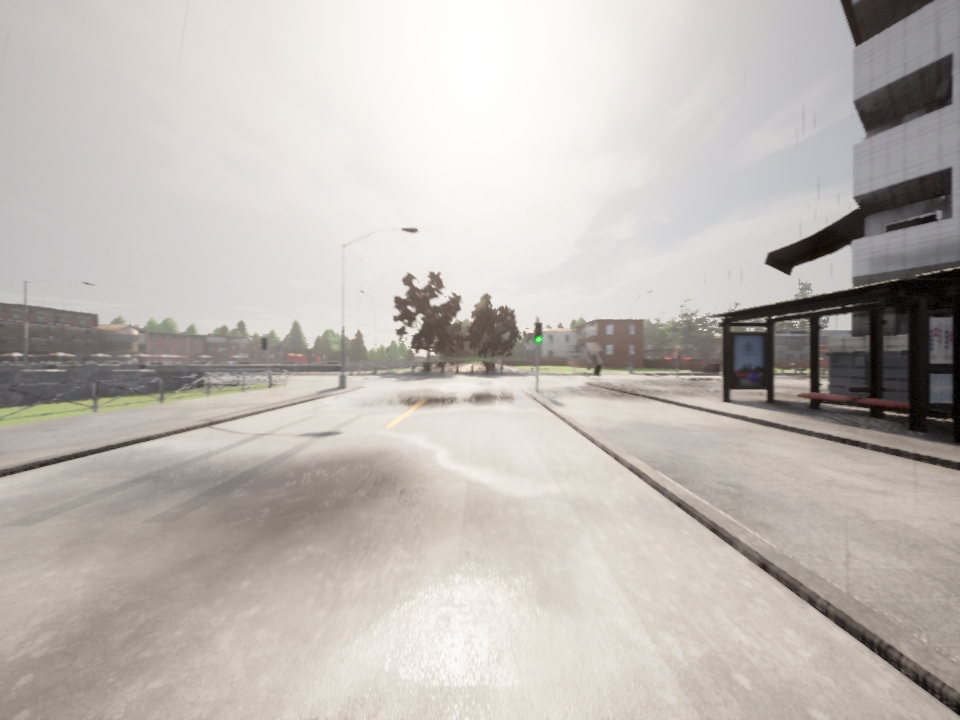} &\hspace{-0.47cm}
\includegraphics[width=0.237\linewidth, height=0.17\linewidth]{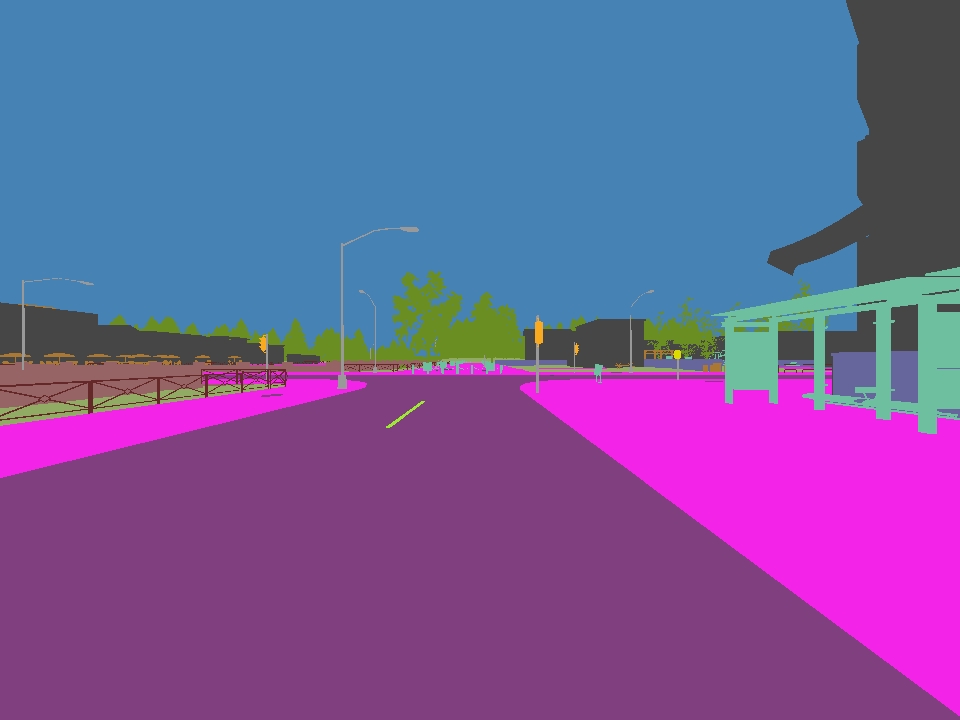} &\hspace{-0.47cm}
\includegraphics[width=0.237\linewidth, height=0.17\linewidth]{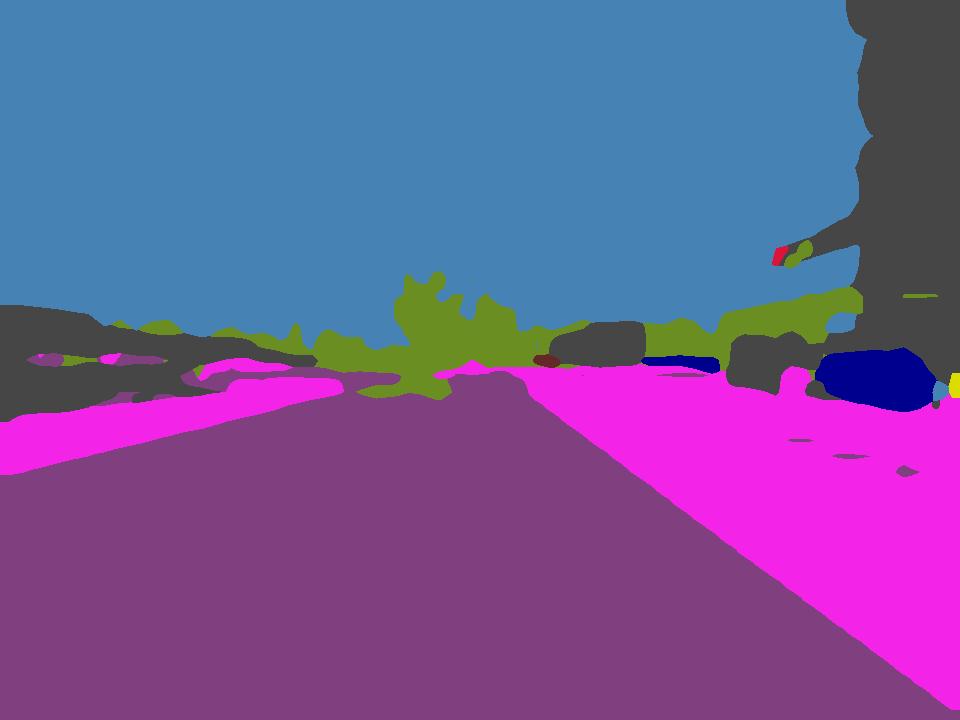} &\hspace{-0.47cm}
\includegraphics[width=0.237\linewidth, height=0.17\linewidth]{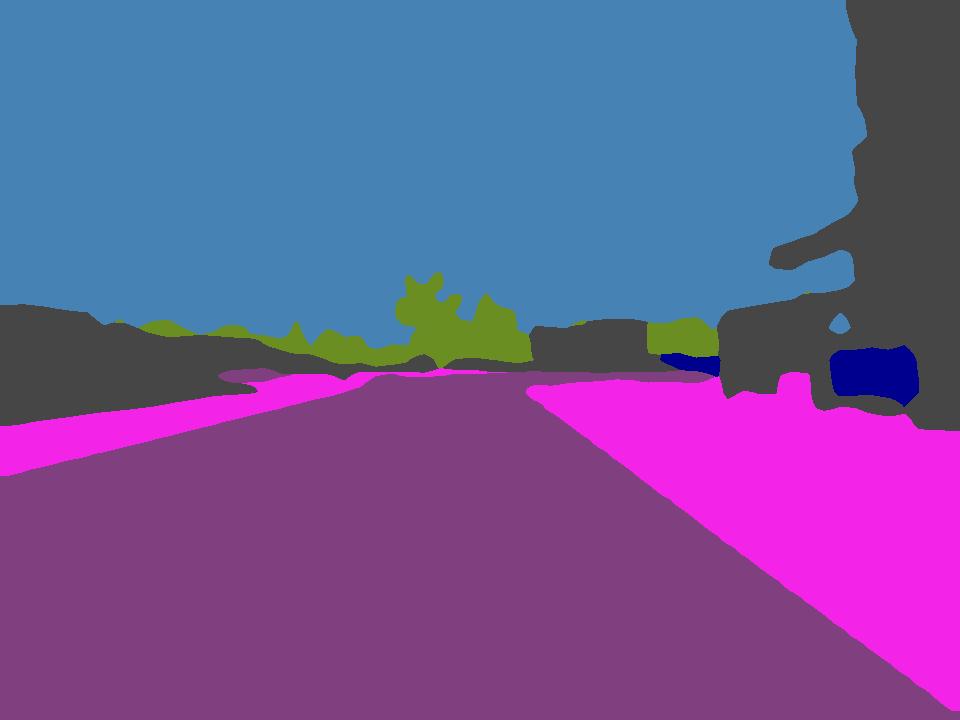}\\
\hline

\includegraphics[width=0.237\linewidth, height=0.17\linewidth]{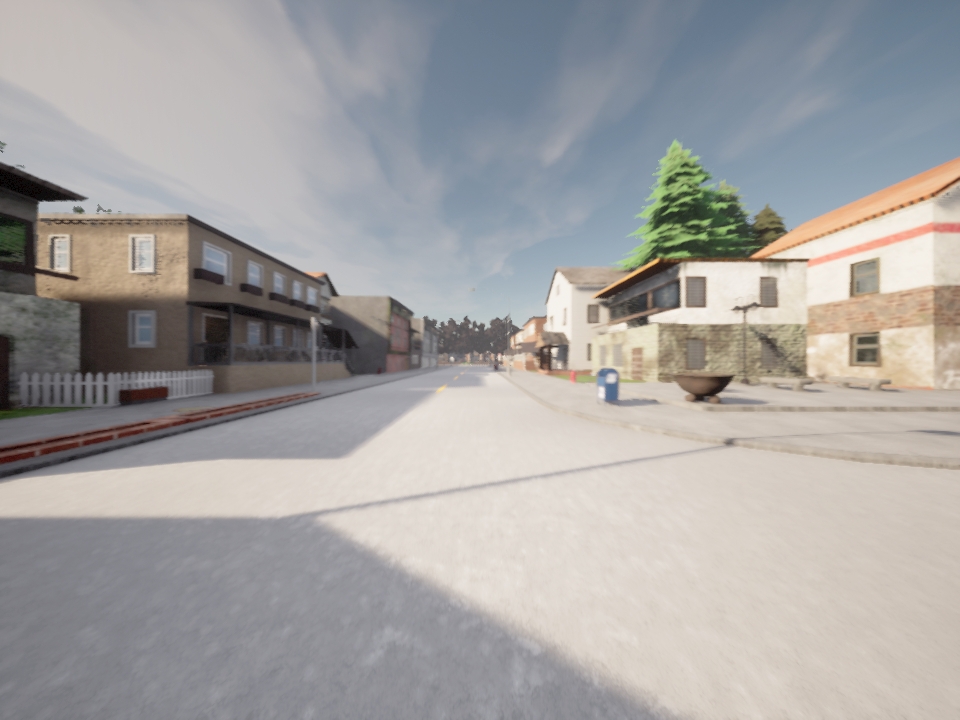} &\hspace{-0.47cm}
\includegraphics[width=0.237\linewidth, height=0.17\linewidth]{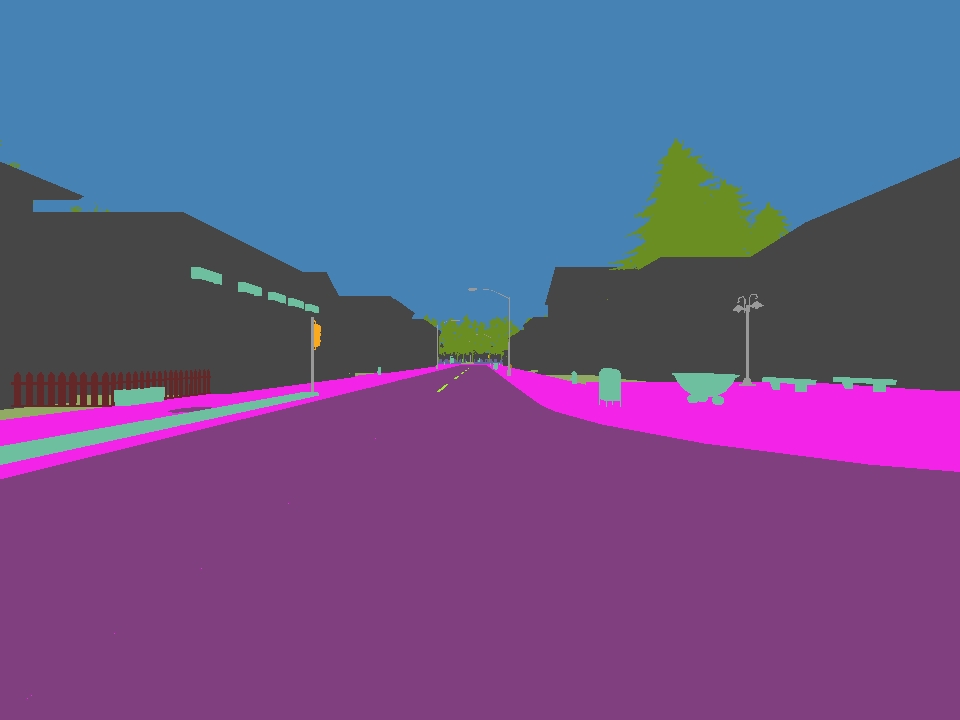} &\hspace{-0.47cm}
\includegraphics[width=0.237\linewidth, height=0.17\linewidth]{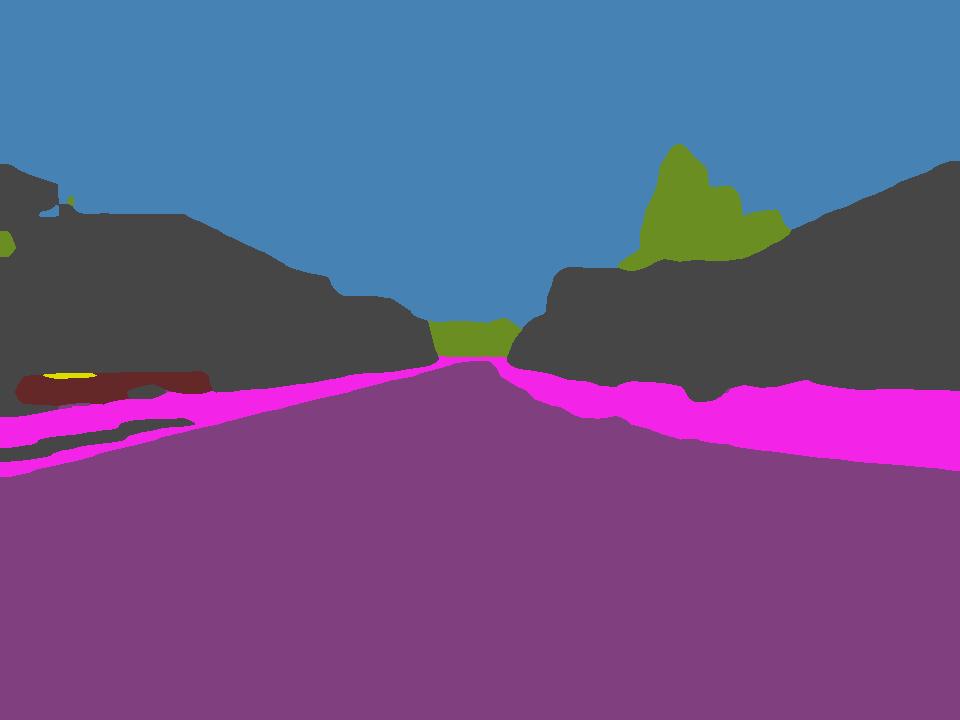} &\hspace{-0.47cm}
\includegraphics[width=0.237\linewidth, height=0.17\linewidth]{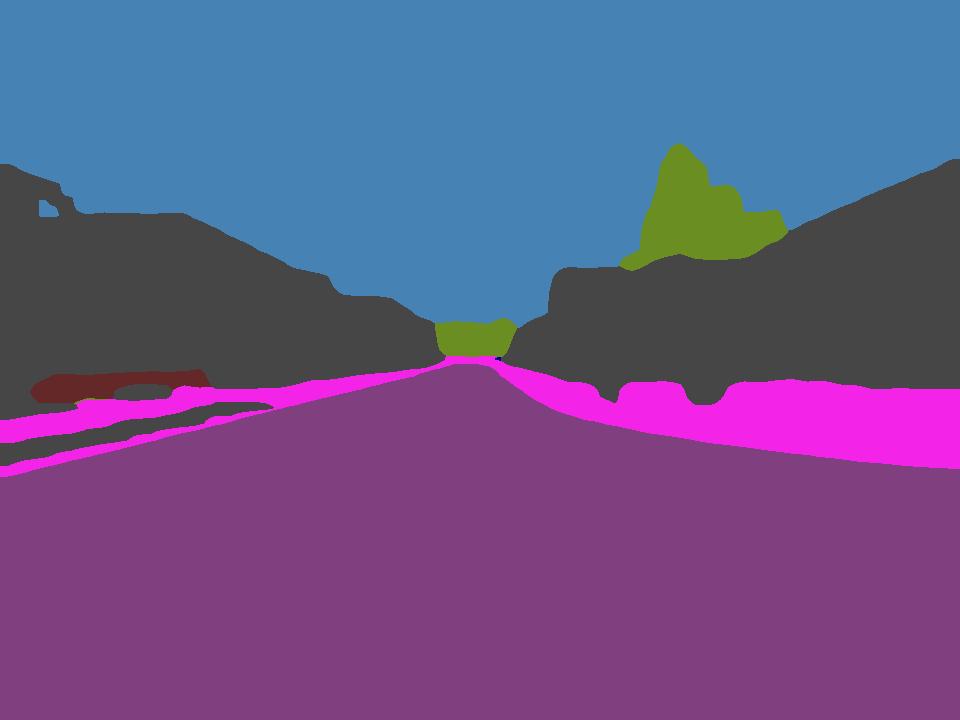}\\
\hline
\end{tabularx}
\label{tab:semantic_pred_carla}
\end{table}

\subsection{Implementation in CARLA World}
\vspace{-0.2cm}
In this section, we implement the proposed FedRC in CARLA simulation world to qualitatively validate our proposed approach. Our methodology involves collecting RGB images, each paired with corresponding semantic tags as depicted in \Cref{Fig.capturing}, which composes the training dataset for the semantic head. Subsequently, upon completing the training phase, we assess the model's performance by comparing the predicted semantic segmentation of previously unseen RGB images from CARLA against the ground truth. This comparison is carried out using the FedRC and FedAvg models. The qualitative outcomes, as presented in \Cref{tab:semantic_pred_carla}, confirm that although some discrepancies in detail against ground truth are observed, the efficacy of the FedRC in AD scenarios is still demonstrated, particularly in the TriSU task.

\section{Conclusion}
In this study, we attempt to improve TriSU model generalization in inter-city setting based on HFL. FedRC is proposed to accelerate HFL TriSU model convergence rate. We conduct comprehensive experiments and compare the results with current state-of-the-art approaches. The findings reveal that FedRC can accelerate HFL TriSU model convergence rate. Future work includes applying FedRC to a wider range of AD tasks and integrating multi-modal data into FedRC.

\end{document}